\documentclass[letterpaper]{article} % DO NOT CHANGE THIS
\usepackage{aaai25}  % DO NOT CHANGE THIS
\usepackage{times}  % DO NOT CHANGE THIS
\usepackage{helvet}  % DO NOT CHANGE THIS
\usepackage{courier}  % DO NOT CHANGE THIS
\usepackage[hyphens]{url}  % DO NOT CHANGE THIS
\usepackage{graphicx} % DO NOT CHANGE THIS
\usepackage{natbib}  % DO NOT CHANGE THIS AND DO NOT ADD ANY OPTIONS TO IT
\usepackage{caption} % DO NOT CHANGE THIS AND DO NOT ADD ANY OPTIONS TO IT
\usepackage{algorithm}
\usepackage[utf8]{inputenc} % allow utf-8 input
\usepackage[T1]{fontenc}    % use 8-bit T1 fonts
\usepackage{url}            % simple URL typesetting
\usepackage{booktabs}       % professional-quality tables
\usepackage{amsfonts}       % blackboard math symbols
\usepackage{nicefrac}       % compact symbols for 1/2, etc.
\usepackage{microtype}      % microtypography
\usepackage{xcolor}         % colors

\RequirePackage{fancyhdr}
\RequirePackage{xcolor} % changed from color to xcolor (2021/11/24)
\RequirePackage{algorithm}
\RequirePackage{eso-pic} % used by \AddToShipoutPicture
\RequirePackage{forloop}
\RequirePackage{url}

\usepackage{amsmath,amsfonts,bm}

\def\eqref#1{equation~\ref{#1}}
\def\1{\bm{1}}

\DeclareMathAlphabet{\mathsfit}{\encodingdefault}{\sfdefault}{m}{sl}
\SetMathAlphabet{\mathsfit}{bold}{\encodingdefault}{\sfdefault}{bx}{n}

\usepackage{url}
\usepackage{scalerel,stackengine}
\stackMath
\newcommand\reallywidehat[1]{%
\savestack{\tmpbox}{\stretchto{%
  \scaleto{%
    \scalerel*[\widthof{\ensuremath{#1}}]{\kern.1pt\mathchar"0362\kern.1pt}%
    {\rule{0ex}{\textheight}}%WIDTH-LIMITED CIRCUMFLEX
  }{\textheight}% 
}{2.4ex}}%
\stackon[-6.9pt]{#1}{\tmpbox}%
}
\usepackage{amsthm}

\usepackage{breqn}
\usepackage{algorithm}
\usepackage{algorithmicx}
\usepackage{algpseudocode}
\usepackage{booktabs}
\usepackage{graphicx}
\usepackage{xcolor}
\usepackage{multirow}
\usepackage{units}
\usepackage{subfig}
\usepackage{pifont}% http://ctan.org/pkg/pifont
\usepackage[capitalize,noabbrev]{cleveref}

\usepackage{xcolor}

\theoremstyle{plain}

\theoremstyle{definition}

\theoremstyle{remark}

\definecolor{my_color}{rgb}{100, 0, 0}
\definecolor{my_color_icml}{rgb}{0, 0, 0}

\usepackage{newfloat}
\usepackage{listings}
\DeclareCaptionStyle{ruled}{labelfont=normalfont,labelsep=colon,strut=off} % DO NOT CHANGE THIS
\floatstyle{ruled}
\newfloat{listing}{tb}{lst}{}
\floatname{listing}{Listing}
\title{Learning Disentangled Equivariant Representation \\ for Explicitly Controllable 3D Molecule Generation}
\author{
    Haoran Liu\textsuperscript{\rm 1, \equalcontrib}, Youzhi Luo\textsuperscript{\rm 1, \equalcontrib}, Tianxiao Li\textsuperscript{\rm 2}, James Caverlee\textsuperscript{\rm 1}, Martin Renqiang Min\textsuperscript{\rm 2}
    \\
}
\affiliations{
    \textsuperscript{\rm 1}Texas A\&M University, College Station, TX 77840;
    \textsuperscript{\rm 2}NEC Laboratories America, Princeton, NJ 08540 \\
    \{liuhr99, yzluo, caverlee\}@tamu.edu, \{tili, renqiang\}@nec-labs.com \\
}

\usepackage{bibentry}
\begin{document}
\maketitle

\begin{abstract}
% Abstracts must be a single paragraph, ideally between 4–6 sentences long. Gross violations will trigger corrections at the camera-ready phase.
% Generating desirable molecular structures in 3D with specific attributes is a crucial aspect of drug discovery. 
We consider the conditional generation of 3D drug-like molecules with \textit{explicit control} over molecular properties 
such as drug-like properties (e.g., Quantitative Estimate of Druglikeness or Synthetic Accessibility score) and effectively binding to specific protein sites. %  is a crucial problem in drug design. 
To tackle this problem, we propose an E(3)-equivariant Wasserstein autoencoder and factorize the latent space of 
our generative model into two disentangled aspects: molecular properties and the remaining structural context of 3D molecules.
Our model ensures explicit control over these molecular attributes while maintaining equivariance of coordinate representation and invariance of data likelihood. 
Furthermore, we introduce a novel alignment-based coordinate loss to adapt equivariant networks for auto-regressive de-novo 3D molecule generation from scratch.
Extensive experiments validate our model's effectiveness on property-guided and context-guided molecule generation, both for de-novo 3D molecule design and structure-based drug discovery against protein targets.

\end{abstract}

\section{Introduction}
\vspace{-5pt}

% Generating desirable molecular structures in 3D is a fundamental problem for drug discovery. 
% Recently, there has been considerable progress in the field of 3D molecule generation using deep learning techniques.

Although a lot of progress has been made in AI-powered 2D molecule design~\cite{bengio2021flow, jin2018junction, shi2019graphaf, xie2021mars}, generating desirable 3D molecules %molecular structures in 3D 
with specific attributes,  e.g., drug-like properties (e.g., QED and SA Score) and the ability to bind with targeted proteins, is still challenging and crucial in drug discovery. 
Given the vastness of the chemical space, estimated to be %on
in the order of $10^{33}$~\cite{polishchuk2013estimation}, conducting an exhaustive search is impractical. %Consequently,
Therefore, our goal is to develop computational methods for %the
direct generation of novel and valid drug-like molecules \textit{conditional} on desired properties. 

%In the context of existing approaches to \textit{conditional} 3D molecule generation, auto-regressive models like G-SchNet~\cite{gebauer2019symmetry} and G-SphereNet~\cite{luo2022an} have been fine-tuned using subsets of favorable data, selected based on specific threshold criteria. However, a limitation of these models is their inability to generate molecules with desired property values.
To %achieve
perform \textit{conditional} 
3D molecule generation, existing auto-regressive models such as G-SchNet~\cite{gebauer2019symmetry} and G-SphereNet~\cite{luo2022an} %have been
can be fine-tuned using subsets of favorable data, selected based on specific threshold criteria. However, these models are limited by their inability to generate molecules with desired property values such as asphericity, QED, SA Score, and binding to target proteins. On the other hand, equivariant diffusion models~\cite{hoogeboom2022equivariant, qiang2023coarse, xu2023geometric} address this limitation by directly incorporating property values as additional inputs. 
Nevertheless, they encounter a %separate
different challenge %as they lack
due to lacking an explicit latent space for %the 
direct manipulation of latent variables, which is required for assigning or anchoring distinct properties to the noise/latent space. Therefore, they %cannot
are unable to explicitly control molecule attributes %, \textit{e.g.}, chemical properties and structure,
in the generation process.
However,
% in many real-world drug design scenarios, specialists need to optimize certain properties while keeping others nearly unchanged. 
% For example, one may want to improve synthetic accessibility of a drug while keep the binding affinity to a target.
explicit control over 3D molecule generation is crucial in real-world drug design scenarios. 
Specialists often need to optimize certain properties while maintaining others. 
For instance, improving a drug's synthetic accessibility without altering its binding affinity to a target. 
This precision demands a generation process with a manipulable latent space, enabling targeted property modifications. % \todo{connection: why we need explicit control and why 3D}
To achieve \textit{explicit control} over 3D molecule generation, we propose to incorporate disentangled representation learning~\cite{bengio2013representation, higgins2016beta, kim2018disentangling} into this task. 
Specifically, we factorize the latent space of our generative model into two disentangled aspects: one representing the desired molecular properties, and the other capturing the remaining structural context related to 3D molecular validity. %and binding interactivity.
% the molecular properties and the remaining component, i.e., the structural context related to 3D molecular validity and binding interactivity. 
% The disentanglement is achieved by two components: a Wasserstein regularization loss~\cite{tolstikhin2018wasserstein} to force the independence between property latent variables and context latent variables, and a prediction head to ensure that the property latent variables contain function-relevant information.
Notably, almost all existing methods only focus on the property-guided side of 3D molecule generation, neglecting the control over the remaining component of the latent space, i.e., the complementary factors to molecular properties.
%Notably, our proposed method ensures explicit control over both molecule attributes and 3D structural context, making it a useful tool in %the 
% innovative and efficient design of pharmaceutical compounds.%, especially for 3D lead optimization. %The
In contrast, our model supports two distinct generation modes: \textbf{property-targeting generation}, for which we generate 3D molecules with specific, desirable properties for property-targeting 3D generation,
%for generating molecules with specific chemical property values or binding functions
and \textbf{context-preserving generation}, for which we generate 3D molecules allowing optimization of targeted molecular properties while maintaining the molecule's fundamental architecture.
% for modifying molecular attributes while preserving overall structure. 
%In drug discovery, generating 3D molecules with specific, desirable properties is a key objective, known as property-targeting 3D generation. 
% Specifically, we  %Additionally, we introduce a novel context-preserving 3D molecule generation task, allowing optimization of targeted molecular properties while maintaining the molecule's fundamental architecture.
% and we generate 3D molecules allowing optimization of targeted molecular properties while maintaining the molecule's fundamental architecture for context-preserving 3D generation. 
%This capability
These two capabilities enabled by our model %is highly relevant to
are essential for many real-world drug design scenarios, where minor modifications can critically impact a drug's efficacy and safety.
% Our model introduces a novel \textit{structural context-preserving 3D generation} task, which enables chemists and researchers to refine targeted molecular properties while preserving the molecule's fundamental architecture. 
% This capability plays a crucial role in enhancing the efficacy, safety, and synthesizability of molecules in drug development. 
%Specifically,
Especially, the \textit{unique capability for context-preserving generation }distinguishes our model from all existing 3D molecule generation methods.
% Specifically, the structure-guided generation task cannot be realized by previous conditional 3D molecule generation methods.
% The former enables generating molecules with specific property values, while the latter focuses on adjusting specific aspects of a molecule's properties while preserving its general structural framework.
 %  while preserving the equivariance of coordinates and invariance of likelihood
% These functionalities of E3WAE, focusing on fine-grained manipulation of molecular generation, are essential tools in the innovative and efficient design of pharmaceutical compounds.

To realize these capabilities, we %further 
propose an \textit{E(3)-equivariant Wasserstein autoencoder} (E3WAE) model architecture %that achieves strong performance in both explicit control and generation performance. The
based on the above
disentangled representation learning framework for controllable 3D molecule generation in this paper. 
The disentanglement is achieved by two components: a Wasserstein regularization loss~\cite{tolstikhin2018wasserstein} to force the independence between property latent variables and context latent variables, and a prediction head to ensure that the property latent variables contain property-relevant information.
Specifically, our model adopts a fragment-based auto-regressive approach, tailored for generating large-scale, drug-like molecules.
However, it is infeasible to directly apply equivariant networks~\cite{batatia2022mace, e3nn_paper, satorras2021n} to auto-regressive 3D molecule generation without any external reference structures. To overcome this challenge, we devise a novel coordinate loss function with structure alignment to guarantee equivariance of coordinate representation and invariance of data likelihood. 
% E3WAE employs two E(3)-equivariant encoders to encode a 3D molecule into the property latent variable and the context latent variable, respectively.  And the latent variables are then combined to reconstruct the molecule in an auto-regressive manner. 
% A prediction head with the supervision of property labels is attached to the property latent variables to ensure they carries information related to the property. 
% In addition, to adapt E(3)-equivariant networks to auto-regressive generation without external reference structure, 
%E3WAE integrates E(3)-equivariant networks, disentangled representation learning and the novel coordinate loss, %achieves performance improvement in property-targeting generation and demonstrates effectiveness in structure-guided generation experiments.
Leveraging E(3)-equivariant networks, disentangled representation learning, and the coordinate loss, E3WAE outperforms previous methods on property-targeting generation and performs effective context-preserving generation on which previous methods fail.

To summarize, our contributions are as follows:
1) We propose %to incorporate 
the \textit{first} disentangled representation learning framework %into
for both \textbf{\textit{de-novo}} and \textbf{strutcure-based} 3D molecule generation% tasks
, achieving \textit{explicit control} over molecule property and structure context.
This benefits real-world drug design scenarios where it is often necessary to fine-tune a specific property while keeping others unchanged.
%To the best of our knowledge, this is the first work on disentangled representation learning for 3D molecule generation.
% 2) We formulate a novel ``context-preserving 3D molecule generation'' task. This capability is highly relevant to many real-world drug design scenarios, where minor modifications can critically impact a drug's efficacy and safety.
2) We %propose a disentangled 
design a novel auto-regressive E(3)-equivariant Wasserstein autoencoder %based framework build upon E(3)-equivariant networks and introduce 
%architecture 
with a new coordinate prediction loss to adapt equivariant networks for \textit{de-novo} 3D molecule generation. %from scratch.
3) We validate our model's effectiveness in property-targeting and the novel context-preserving generation tasks for both \textit{de-novo} 3D molecule generation and structure-based drug design. 
% 3) We validate our model's effectiveness in property-targeting and context-preserving generation on both \textit{de-novo} 3D molecule generation and structure-based drug design tasks. -> Practical use in drug design applications
% Visualizations also confirm latent space disentanglement via distinct property and mixed structural clusters aligned to property labels.

% \paragraph{Main Contribution}
% \begin{itemize}
%     \item Disentangled representation learning for 3D molecules
%     \item Equivariant Auto-regressive model for \textit{de-novo} molecule generation (Coordinate loss)
%     \item experiments
% \end{itemize}

% \todo{Define explicit control}
% "Explicit control" refers to the ability to directly manipulate or specify certain attributes of the models' outputs. This means the model is designed in a way that allows for precise adjustments or settings of specific features or properties of the generated items. For instance, in molecule generation, explicit control could involve adjusting the shape, size, functional groups, or other chemical properties of a molecule. This capability is crucial in areas like drug design or materials science, where fine-tuning these characteristics can directly impact the effectiveness or functionality of the resulting compounds or materials.

% \vspace{-3pt}
\section{Related Works}
\label{sec:related_works}
\vspace{-5pt}
\textbf{De-novo Molecular Generation in 3D. } 
% Equivariance vs invariance; \\
% Atom vs Fragment-based Generation. \\
% Pitfalls: none of them emphasize on conditional generation. \\
Recently, there has been considerable progress in the field of 3D molecule generation using deep learning techniques. Various approaches have been explored to accurately represent and generate 3D molecular structures. Some have focused on learning joint distributions of geometric features~\cite{garcia2021n, luo2022an}. Others, such as G-SchNet~\cite{gebauer2019symmetry}, have employed auto-regressive models with symmetry constraints to sample 3D molecules directly in their spatial configurations. A notable advancement is the use of equivariant diffusion models~\cite{hoogeboom2022equivariant, qiang2023coarse, xu2023geometric}, which have shown significant promise in generating high-quality molecules. These techniques have been further applied in specific generation tasks with external conditioning, such as reference ligands~\cite{adams2023equivariant, liu2022generating, powers2022fragmentbased, guan2023d} and linkers~\cite{ huang20223dlinker, igashov2022equivariant, imrie2020deep}. 
Despite these advancements, a common limitation in most existing models is the lack of \textit{explicit control} over the attributes of the generated molecules. 
This poses a challenge for applications where there is \textit{a need to modify certain attributes of the molecules}, such as a specific chemical property, \textit{while preserving other aspects} like overall shape and composition. Our research aims to address this gap by \textit{explicitly controlling} the generation of molecules, where higher-level semantics can be precisely manipulated to meet specific criteria.

\textbf{Disentangled Representation Learning.}
The design of disentangled representation learning (DRL) tasks varies based on the nature of data involved.  For instance, DRL tasks involve decoupling ``style'' and ``content'' in text~\cite{cheng2020improving} or images~\cite{kotovenko2019content}, or distinguishing between ``static'' and ``dynamic'' components in videos~\cite{zhu2020s3vae, han2021disentangled}. Such disentangled representations are pivotal in applications like style transfer~\cite{cheng2020improving, kotovenko2019content, lee2018diverse} and conditional generation~\cite{denton2017unsupervised, zhu2018visual}. To achieve disentanglement, many regularization techniques are developed. These techniques include managing the capacity of latent space~\cite{burgess2018understanding, higgins2016beta}, implementing adversarial loss~\cite{deng2020residual}, employing cyclic reconstruction~\cite{lee2018diverse}, imposing mutual information constraints~\cite{chen2016infogan, cheng2020improving, zhu2020s3vae}, and integrating self-supervising auxiliary tasks~\cite{zhu2020s3vae}. 
In molecular science, DRL has been applied to molecule generation using SMILES sequences~\cite{mollaysa2020conditional} and 2D molecular graphs~\cite{du2022small}. However, to the best of our knowledge, its application in \textit{de-novo} 3D molecule generation and structure-based drug design has not been explored.

% \todo{Preliminary on Equivariance and Invariance}

\vspace{-5pt}
\section{Preliminaries}
\vspace{-5pt}
\textbf{Problem Formulation.} 
To generate \textit{large-scale drug-like} molecules, our method adopts an auto-regressive, fragment-based 3D generation approach. 
% This approach significantly reduces computational complexity while accurately modeling the chemical and geometric properties of molecules.
Let $\mathcal{G}$ be the space of 3D molecular graphs, where each 3D molecular graph $G\in \mathcal{G}$ consists of the fragment node set $\mathcal{V}$, the edge set $\mathcal{E}$, and the fragment coordinate matrix $\mathcal{R}$. 
Specifically, each fragment represents a combination of several atoms and bonds, e.g., a benzene ring could be a fragment that includes six carbon atoms and aromatic bonds. Correspondingly, each fragment $i \in \mathcal{V}$ is associated with a node feature $v_i$ and a position vector $\mathbf{r}_i$, which corresponds to the $i$-th line of $\mathcal{R}$ and represents the center coordinates of the fragment.
Following %previous works on 
fragment-based generation~\cite{bengio2021flow, jin2018junction, qiang2023coarse}, we use edge $e_{ij} \in \mathcal{E}$ to indicate that two fragments $i, j \in \mathcal{V}$ share a bond/atom.
This fragmentization inherently includes assembled rules for neighboring fragments and enables us to model a substantial portion of the chemical space with a reasonably sized fragment vocabulary.
The generation process is iterative; at each step $t$, the next graph state is predicted as $\mathcal{G}_{t+1} = \mathcal{M}\left(\mathcal{G}_{t} \right)$, where $\mathcal{M}$ is the auto-regressive generation model.
The fragmentization details are in Appendix~\ref{appendix: fragment}.
% Notably, we design our model to have explicit control over molecule attributes and maintain E(3)-invariance and equivariance properties during generation steps.
Building upon this, our model is designed for explicit control over molecular attributes while ensuring both E(3)-invariance and equivariance properties.
% We take an auto-regressive approach to generate the 3D molecule fragment-by-fragment. Concretely, in each iteration t, the graph for next iteration is predict as $\mathcal{G}_{t+1} = \mathcal{M}\left(\mathcal{G}_{t} \right)$, where $\mathcal{M}$ is the auto-regressive generation model.
% We design our model to have explicit control over molecule attributes and maintain E(3)-invariance and equivariance properties during generation steps.

% This kind of definition enables us to model molecule geometry using tangent condition on fragment sphere, with a reasonable size fragment vocabulary. 
% Therefore, the target of a 3D generation model is to learn a probabilistic model $P_\theta(V, E)$ to model the empirical distribution of 3D molecule graphs, modeling the underlying physical and chemical constraints, which could also be used to sample new molecules from the chemical space.

\begin{figure*}[tp]
    \vspace{-10 pt}
    \centering
    \includegraphics[width=0.75\textwidth]{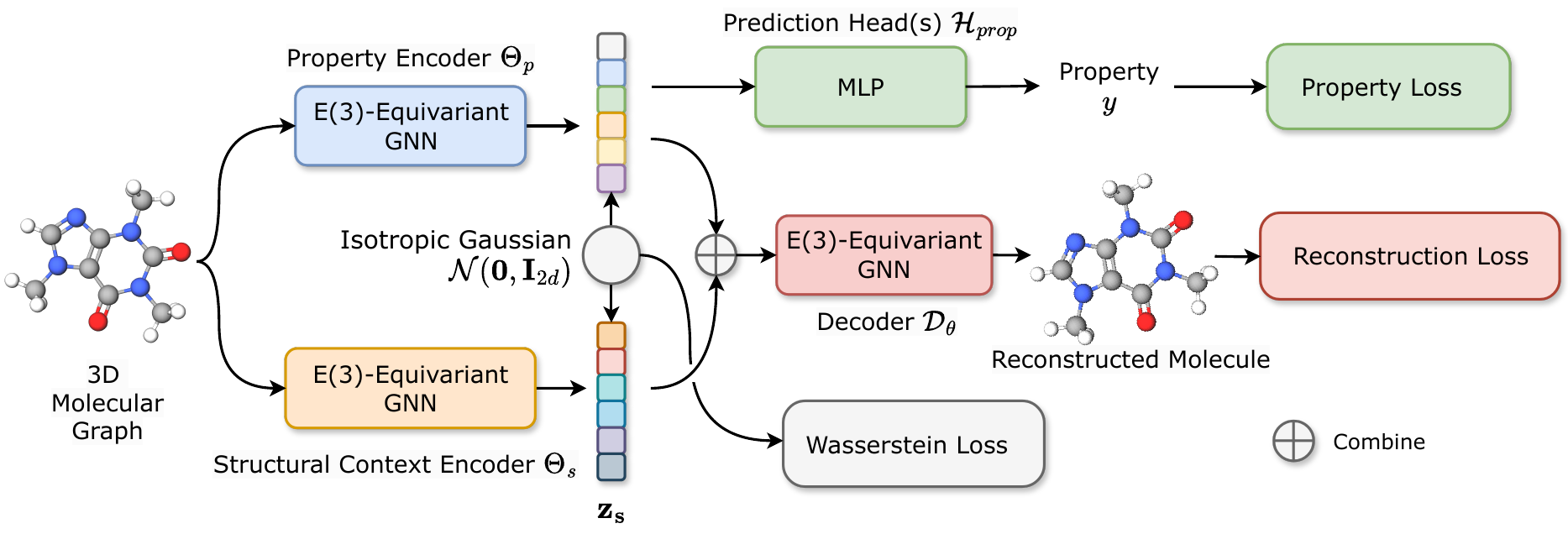}
    \vspace{-10 pt}
    \caption{An illustration of the proposed E3WAE framework. A 3D molecule is encoded into two disentangled latent variables: the property variable $\mathbf{z}_p$ and the structural context variable $\mathbf{z}_s$ with two E(3)-equivariant encoders, respectively.  The latent variables are then combined to reconstruct the molecule in an auto-regressive manner. A prediction head with the supervision of property labels is attached to $\mathbf{z}_p$ to ensure $\mathbf{z}_p$ carries information related to the property. 
    Disentanglement of the property and structure variables is achieved through a Wasserstein autoencoder regularization loss and minimization of the MMD distance against an isotropic Gaussian distribution. The overall training objective combines a reconstruction loss, a property prediction loss, and the Wasserstein loss. 
    }
    \label{fig:framework}
    \vspace{-10 pt}
\end{figure*}

\textbf{Explicit Control. } 
We define a generative model with explicit control as $\mathcal{M}: \mathcal{Y}_1 \times \mathcal{Y}_2 \rightarrow \mathcal{X}$, where $\mathcal{Y}_1$ and $\mathcal{Y}_2$ represent the attribute spaces of two different attributes, and $\mathcal{X}$ is the space of generated objects. The model $\mathcal{M}$ is considered to have explicit control if, for any given attributes $y_1 \in \mathcal{Y}_1$ and $y_2 \in \mathcal{Y}_2$, it consistently generates an object $x \in \mathcal{X}$ reflecting these attributes. In our model, $\mathcal{Y}_1$ is the space of target molecular properties, and $\mathcal{Y}_2$ refers to the structural context of molecules. Manipulating $y_1$ allows for altering molecular properties and maintaining their structure context, whereas adjusting $y_2$ enables refining molecule structures and maintaining their chemical properties.

% \textbf{Explicit Control.} Formally, we define a generative model with explicit control as $\mathcal{M}: \mathcal{Y}_1 \times \mathcal{Y}_2 \rightarrow \mathcal{X}$, where $\mathcal{Y}_1$ and $\mathcal{Y}_2$ are the spaces of specific attributes, and $\mathcal{X}$ is the space of generated objects. 
% In this work, we define $\mathcal{Y}_1$ represents the space of target property attributes, $\mathcal{Y}_2$ is the space of structure context of the molecules
% The model $\mathcal{M}$ is considered to have explicit control if, for any given attribute $y_1 \in \mathcal{Y}_1$ and $y_2 \in \mathcal{Y}_2$, the model generates an object $x \in \mathcal{X}$ that reflects the specific attribute $y_1$ and $y_2$. By manipulating $y_1$, we enable property-targeting generation of molecules, tailoring specific properties as desired. 
% Simultaneously, by adjusting $y_2$, we enable structure-guided generation, making it possible to refine specific aspects of a molecule's properties while preserving its general structural framework.

\textbf{E(3)-Invariance and Equivariance.} 
$\text{E}(3)$ or 3D Euclidean group is the group generated by all 3D rotations, translations and reflections in the 3D space.
In the context of \textit{de-novo} 3D molecule generation from scratch, i.e. without external reference structure, the goal is to design models maintaining equivariance of coordinate representation and invariance of data likelihood.
First, the coordinates of a molecule's nodes are \textit{equivariant} to the positions of other nodes in the 3D molecule. 
That means, when generating the coordinates $r_t$ for an atom at iteration $t$, if we rotate or translate the current input structure $\mathcal{R}_t$, then $r_i$ needs to be rotated or translated correspondingly.
Formally, consider a rotation matrix $\mathbf{R} \in SO(3)$ and a translation vector $\tau \in \mathbb{R}^3$, for the coordinate generation model $\mathcal{M}^r$, we have $\mathbf{R} r_t + \tau = \mathcal{M}^r(\mathbf{R}\mathcal{R}_t + \tau)$.
In contrast, the likelihood $\pi(r_i |\mathcal{R}_t)$ should be \textit{invariant} to rotations and translations as
they do not change the 3D structure, so that $\pi(\mathbf{R} r_t + \tau | \mathbf{R}\mathcal{R}_t + \tau) = \pi(r_t |\mathcal{R}_t)$, for any rotation $\mathbf{R} \in \mathbb{R}^{3\times 3}$ or translation $\tau \in \mathbb{R}^3$.
% The equivariance and invariance properties ensures realistic and consistent structure generation, irrespective of spatial orientations and positions.
% \todo{Check correspondence with Section~\ref{sec:method} and ~\ref{sec: loss}.}
% Group is a set of operations equipped with multiplications, associativity, the identity element and the inverse element.
% The group of all 3D rotations, translations and reflections is called 3D Euclidean Group or $\text{E}(3)$.
% Consider a function $\phi: X \rightarrow Y$, where $X$ is the input space and $Y$ is the output space. The function $\phi$ is equivariant to $\text{E}(3)$ if for every element $g \in \text{E}(3)$ and every $x \in X$, the following holds: $\phi(gx) = g\left(\phi(x)\right)$. This means that changes in the input, such as rotating or translating a molecule, are consistently reflected in the output. In contrast, E(3)-invariance implies $\phi(gx) = \phi(x)$ for all $g \in \text{E}(3)$, signifying the output remains unchanged irrespective of the spatial transformations applied to the input. 
% where the properties of generated molecules are \textit{invariant} (unchanged by spatial transformations), while the coordinates of a molecule's nodes are \textit{equivariant} to the positions of other nodes in the 3D molecule. 
% For \textit{de-novo} 3D molecule generation tasks, the generated molecules should be invariant to the scalar properties and equivariant to all other nodes' coordinates.

\vspace{-5pt}
\section{Methodology}
\label{sec:method}

%% Disentangle ACL, CVPR 2020 S3VAE, ICLR

% \subsection{Problem Formulation}

% We propose an \textit{E(3)-equivariant Wasserstein autoencoder} (E3WAE) model, where the 3D molecule is generated from the latent variable $\mathbf{z}$ and $\mathbf{z}$ is factorized into two disentangled variables: the property variable $\mathbf{z}_p$ and the structure variable $\mathbf{z}_s$. The former variable comprises the chemical property of the 3D molecule while the later refers to all other patterns that do not relate to the property yet reflecting the chemical constraints within the molecular chemical space.
% This factorization allows for \textit{explicit control} during the generation process. 
% This bifurcation of the latent space grants precise control during the generative process. 
% By manipulating $\mathbf{z}_p$, we enable property-targeting generation of molecules, tailoring specific properties as desired. 
% Simultaneously, by adjusting $\mathbf{z}_s$, we enable structure-guided generation, making it possible to refine specific aspects of a molecule's properties while preserving its general structural framework.

We propose an \textit{E(3)-equivariant Wasserstein autoencoder} (E3WAE) model, where the generation of the 3D molecule is factorized into two disentangled factors: the \textbf{property} and the \textbf{structural context}. 
The former variable comprises the chemical property of the 3D molecule while the latter refers to all other 3D structure patterns that do not relate to the property yet reflect the chemical constraints within the molecular chemical space.
This factorization allows for \textit{explicit control} during the generation process. 
In addition, we introduce a novel coordinate prediction loss %(see Section~\ref{sec: loss}) 
as part of the reconstruction loss to enable auto-regressive 3D molecule generation with equivariant networks. 
An overview of our model framework is shown in Figure~\ref{fig:framework}.
% Architecture detail on Arxiv.

% \todo{where to put this def. }
% Formally, we define a generative model with explicit control as $\mathcal{M}: \mathcal{Y} \times \mathcal{Z} \rightarrow \mathcal{X}$, where $\mathcal{Y}$ represents the space of control attributes, $\mathcal{Z}$ is the latent variable space, and $\mathcal{X}$ is the space of generated objects. The model $\mathcal{M}$ is considered to have explicit control if, for any given attribute $y \in \mathcal{Y}$ and any latent variable $z \in \mathcal{Z}$, the model generates an object $x \in \mathcal{X}$ that reflects the specific attribute $y$. 

% \subsection{E(3) Equivariant Disentangled Autoencoder}
\subsection{E(3)-Equivariant Disentangled Encoder}
\label{sec: encoder}
% VN-MLP~\cite{deng2021vector}
We use two separate E(3)-equivariant GNN encoders to extract invariant and equivariant latent variables, respectively:
\begin{align}
    \mathbf{z}_{h, e}, \, \mathbf{z}_{v, e} = \Theta_e(\mathcal{G}), 
\end{align}
where $e = \{p, s\}$ represents ``property'' or ``structural context''. The invariant latent variable associate with node $i$, $z_{h, e}^i \in \mathbf{z}_{h, e}$, has a dimensionality of $d_h$. Correspondingly, the equivariant latent variable $\mathbf{z}_{v, e}^i \in \mathbf{z}_{v, e}$ is dimensioned in $\mathbb{R}^{d_v \times 3}$.

In this paper, we opt for E(3)-equivariant Vector Neurons Multilayer Perceptron (VN-MLP)~\cite{deng2021vector} and Mixed-Features Message Passing (MF-MP)~\cite{huang20223dlinker} as building blocks for both branches of encoders $\Theta_e$.
These choices are due to their proven effectiveness in integrating both invariant and equivariant features, ensuring the property space's invariance and the coordinate space's equivariance. 
%due to their effectiveness in combining invariant and equivariant features. 
However, it is important to note that our model is compatible with any E(3)-equivariant architecture.

% \subsection{Auxiliary Property Prediction Head}
For the \textit{property} branch, the property latent variables $\mathbf{z}_{h, p}$ and $\mathbf{z}_{v, p}$ are first derived by the property encoder $\Theta_p(G)$.
Subsequently, a Readout function is used to obtain graph-level representations. The Readout function is implemented as either the average or the summation of all node embeddings.
In order to ensure that these latent variables carry information related to the property of 3D molecules, we introduce an auxiliary prediction head $\mathcal{H}_{prop}$ that takes the aggregated $\mathbf{z}_{h, p}$ as input and predicts the target property value of the 3D molecule:
\begin{align}
    \hat{y} = \mathcal{H}_{prop}(\text{Readout}(\mathbf{z}_{h, p})),
\end{align}

For the \textit{structure context} branch, the context latent variables $\mathbf{z}_{h, s}$ and $\mathbf{z}_{v, s}$ are encoded using the structural encoder $\Theta_s(G)$. 
To ensure that these variables capture comprehensive information about the molecule, specifically excluding patterns related to the target property, we employ an auto-regressive molecule reconstruction loss. 
Details of this process are further discussed in Section~\ref{sec: decoder}.

\subsection{Disentanglement of the Latent Space}
\label{sec: disentangle}

To achieve the disentanglement between property and context latent variables, we introduce a Wasserstein autoencoder regularization loss following~\citet{tolstikhin2018wasserstein}.   
This approach involves minimizing the Maximum Mean Discrepancy (MMD) between the distribution of latent variables and an isotropic multivariate Gaussian prior, denoted as
$\mathbf{z} \sim P_{\mathbf{z}}$.
Specifically, for the invariant latent variables 
$\mathbf{z}_h \sim Q_{\mathbf{z}}^h$, where $\mathbf{z}_h = \text{concat}(\mathbf{z}_{h,p}, \mathbf{z}_{h,s})$, and an isotropic Gaussian prior
$P_{\mathbf{z}}^h = \mathcal{N} \left(\mathbf{0}, \mathbf{I}_{2 d_h}\right)$, 
we compute the disentanglement loss for invariant variables as
\begin{align}
    \mathcal{L}_{\text{Dis}}^h = \text{MMD}(P_{\mathbf{z}}^h, Q_{\mathbf{z}}^h).
\end{align}
For the equivariant latent variables, $\mathbf{z}_v = \text{concat}(\mathbf{z}_{v,p}, \mathbf{z}_{v,s})$ and $\mathbf{z}_v \in \mathbb{R}^{2d_v \times 3}$, we maintain independence along the $2d_v$ axis while allowing for covariance along the remaining dimension. 
To this end, we sample \textit{three} isotropic Gaussian priors from $P_{\mathbf{z}}^v = \mathcal{N}(\mathbf{0}, \mathbf{I}_{2 d_v})$ and calculate the corresponding disentanglement Wasserstein loss as:
\begin{align}
    \mathcal{L}_{\text{Dis}}^v = \text{MMD}(\mathbf{P}_{\mathbf{z}}^v, \mathbf{Q}_{\mathbf{z}}^v),
\end{align}
where $\mathbf{P}_{\mathbf{z}}^v \in \mathbb{R}^{2d_v \times 3}$ is the combined Gaussian priors and $\mathbf{Q}_{\mathbf{z}}^v \in \mathbb{R}^{2d_v \times 3}$ is the distribution of the equivariant latent variables $\mathbf{z}_v$. 
The total disentanglement loss is then calculated as the sum of these components: $\mathcal{L}_{\text{Dis}} = \mathcal{L}_{\text{Dis}}^h + \mathcal{L}_{\text{Dis}}^v$.

For MMD estimation, we take an input batch of latent variables $\{\mathbf{z}_i\}_{i=1, \cdots, m}$ with batch size $m$. Corresponding samples $\{\mathbf{\tilde{z}}_i\}_{i=1, \cdots, m}$ are randomly drawn from the Gaussian prior, maintaining the same sample size. The MMD is then calculated using the linear time unbiased estimator
~\cite{gretton2012kernel, sutherland2016generative} as
\begin{equation}
\begin{aligned}
    & \text{MMD} (P_{\mathbf{z}}, Q_{\mathbf{z}}) = \frac{1}{\lfloor m/2 \rfloor} \sum_{i=1}^{\lfloor m/2 \rfloor} \left[ k(\mathbf{z}_{2i-1}, \mathbf{z}_{2i}) \right. \\ 
    &\left. 
    + k(\mathbf{\tilde{z}}_{2i-1}, \mathbf{\tilde{z}}_{2i}) - k(\mathbf{z}_{2i-1}, \mathbf{\tilde{z}}_{2i}) - k(\mathbf{z}_{2i}, \mathbf{\tilde{z}}_{2i-1}) \right].
    \label{eq: mmd}
\end{aligned}
\end{equation}
where $k$ is the kernel function, implemented using a radial basis function (RBF) kernel with $\sigma = 1$. 

Minimizing this loss aligns the joint distribution of the latent embeddings with the isotropic normal distributions, so that the property and context latent variables are independent. 
Additionally, the Gaussian shape of the latent space facilitates smooth interpolation, effective regularization, and enhanced generation diversity~\cite{kingma2013auto, tolstikhin2018wasserstein}.
A theoretical analysis supporting the disentanglement guarantee is provided in Appendix~\ref{appendix: disentanglement_guarantee}.

% \begin{figure}[tp]
%     % \vspace{-10 pt}
%     \centering
%     \includegraphics[width=0.45\textwidth]{figs/reconstruction-1.pdf}
%     \vspace{-10 pt}
%     \caption{An illustration of the auto-regressive reconstruction process. Color indicates different fragment types. }
%     \label{fig:reconstruct}
%     \vspace{-5 pt}
% \end{figure}

\subsection{Decoder and 3D Molecule Graph Reconstruction}
\label{sec: decoder}

The decoder $\mathcal{D}(\mathcal{G}|\mathbf{z}_{h}, \, \mathbf{z}_{v})$ reconstructs 3D molecule graph $\mathcal{G}$ fragment-by-fragment in an auto-regressive manner. Specifically, we employ another E(3)-equivariant GNN as the decoder. We incorporate the connectivity rules between fragments by masking out invalid edges.
For the auto-regressive reconstruction, our approach is inspired by the \textit{focus and expand} molecule generation procedure in previous works~\cite{ceylan2018conditional, huang20223dlinker}. 
Notably, our method is significantly different from previous works.
% in its novel approach to \textit{de novo} molecule generation. 
% Unlike these works, w
We use fragments as nodes and introduce a novel coordinate prediction loss (see Section~\ref{sec: loss}) specifically designed to optimize equivariant networks. 
This approach enables the generation of molecules using equivariant networks without relying on any external conditioning, such as reference structures like linker designs and pocket-based generation.
A detailed algorithm for this decoding process is in Appendix~\ref{appendix: decode_alg}.
Specifically, the decoding process includes the following steps:
(\romannumeral1) \textbf{Node type prediction}: 
Initially, fragment types $\{x_i\}_{i=1}^{n}$ for all $n$ nodes in $\mathcal{G}$ are determined from latent variables $\mathbf{z}_{h}$ and $\mathbf{z}_{v}$. 
The fragment type logits are obtained from latent variables through a self-attention mechanism and an MLP, and these logits are then used to sample all fragment types.  Once the node types are sampled, their embeddings are concatenated with the corresponding latent variables $\mathbf{z}_{h}^i$ for subsequent steps.
(\romannumeral2) \textbf{Focus queue initialization}: 
A focus queue $Q$ is then initialized with a randomly chosen node from $\mathcal{G}$.
(\romannumeral3) \textbf{Focus and expand iterations}: 
For each \textit{focus} node $f$ popped from $Q$, the decoder predicts an \textit{expand} edge connecting $f$ to another node. 
If the connected node $u$ is not a stop node or being linked for the first time, its coordinate $\left(x_u, y_u, z_u\right)$ is predicted. 
Node $u$ is then added to $Q$ if unvisited. 
This process repeats until a stop node is reached, marking $f$ as \textit{visited}. 
The order of node focus and edge connection is determined using Breadth-First Search, facilitating teacher-forcing training. 
The reconstruction process ends when $Q$ is empty, ensuring that all nodes in the connected component of $\mathcal{G}$ have been considered for expansion. 

Concretely, in each iteration $t$ of the focus and expand phase, we have the currently reconstructed subgraph $\mathcal{G}_t = (\mathcal{V}_t, \mathcal{E}_t, \mathcal{R}_t)$. Initially, the latent node embeddings are updated as $\hat{\mathbf{z}}_{h}, \hat{\mathbf{z}}_{v}$ with a MF-MP layer. Then the edge logits between the focus node $f$ and any node $i$ are obtained with
\begin{align}
e_{f,i} = \Phi\left(\hat{\mathbf{z}}_{h}^f, \hat{\mathbf{z}}_{h}^i, \|\hat{\mathbf{z}}_{v}^f\|, \|\hat{\mathbf{z}}_{v}^i\|, \text{m}_{f, i}, \sum_{j\in\mathcal{V}_t} \hat{\mathbf{z}}_{h}^j, \sum_{j\in\mathcal{V}_t} \hat{\mathbf{z}}_{h}^j \right),
\end{align}
where $\Phi$ is a feed forward network and $\text{m}_{f, i} \in \{0, 1\}$ indicates whether fragments $f$ and $i$ can be connected, based on feasible chemical bonding rules. The softmax function is applied to the edge logits to determine the probabilities for each edge. Node $u$ is then determined by identifying which node $i$ has the highest probability.

To predict the coordinate for the newly connected node $u$, we take the geometric center $\bar{\mathbf{r}}_t$ of current subgraph $\mathcal{G}_t$ as a reference point and predict a displacement of node $u$ related to the reference point.
Initially, we compute two sets of pair-wise interactions in the current connected graph as
\begin{align}
    p_{ij}^{(k)} = \Psi(\hat{\mathbf{z}}_{h}^i, \, \hat{\mathbf{z}}_{h}^j, \, 	
    \langle \mathbf{W}^{(k)} \hat{\mathbf{z}}_{v}^i, \,  
    \mathbf{W}^{(k)} \hat{\mathbf{z}}_{v}^j\rangle), \quad k=1,2
\end{align}
where $\Psi$ is a feed forward network and $\mathbf{W}^{(k)}$ is a learnable linear transformation.
The displacement of node $u$ is then calculated by: 
\begin{align}
    \mathbf{d}_u = \sum_{j\in\mathcal{V}_t} p_{uj}^{(1)}(\mathbf{r}_j - \bar{\mathbf{r}}_t)
    + \Omega_1\left(\sum_{j\in\mathcal{V}_t} p_{uj}^{(2)} \Omega_2(\hat{\mathbf{z}}_{v}^u, \hat{\mathbf{z}}_{v}^j)\right)
\end{align}
where $\Omega_1$ and $\Omega_2$ are VN-MLP layers. 
Finally, the predicted coordinates of node $u$ is obtained by $\hat{\mathbf{r}}_u = \mathbf{d}_u + \bar{\mathbf{r}}_t$.

\subsection{Training Objectives}
\label{sec: loss}

The proposed E3WAE network is trained by minimizing a weighted sum of three individual losses:
\begin{align}
    \mathcal{L}_{\text{Total}} = \mathcal{L}_{\text{Prop}} + \alpha \mathcal{L}_{\text{Dis}} + \beta \mathcal{L}_{\text{Recon}}, 
\end{align}
where $\alpha$ and $\beta$ are the trade-off weights for the losses. Specifically, $\mathcal{L}_{\text{Dis}}$ is introduced in Section~\ref{sec: disentangle}.
In addition, for the auxiliary property prediction head $\mathcal{H}_{prop}$ attached to property encoder $\Theta_p$ in Section~\ref{sec: encoder}, 
% Given that this task is of a regression nature, 
we adopt the L1 loss as the property prediction loss:
\begin{align}
    \mathcal{L}_{\text{Prop}} = \|y - \hat{y}\|_1,
\end{align}
where $y$ denotes the ground truth value of the target property. Finally, for the decoder in Section~\ref{sec: decoder}, we adopt a reconstruction loss which consists of three parts:
\begin{align}
    \mathcal{L}_{\text{Recon}} = \mathcal{L}_{\text{NodeType}} + \mathcal{L}_{\text{Edge}} + \mathcal{L}_{\text{Coords}}.
\end{align}
Initially, a cross-entropy loss $\mathcal{L}_{\text{NodeType}}$ is used for a classification task to accurately determine the types of nodes. Following this, another cross-entropy loss, $ \mathcal{L}_{\text{Edge}} $, is applied to predict the edges in each iteration of the process. 
Finally, we incorporate a log-MSE loss~\cite{yu2020tutorial} as the coordinate prediction loss $\mathcal{L}_{\text{Coords}}^t$ at each iteration $t$:
%, to measure the accuracy of the position prediction for each node. 
\begin{align}
    \mathcal{L}_{\text{Coords}}^t = %\frac{1}{\sum_{i=1}^m f_i}
    \nicefrac{\log \left( \sum_{i=1}^m f_i \, \|\tilde{\mathbf{r}}_{i} - \mathbf{r}_{i}\|^2\right) }{ \left(\sum_{i=1}^m f_i \right)},
\end{align}
where $f_i$ is a binary flag indicating the presence of a newly added node in the $i$-th subgraph $\mathcal{G}_t^{(i)}$ that is not a stop node.  
%indicating whether the predict node is the node to link for the current subgraph $\mathcal{G}_t$. 
% there's new add node connected to the subgraph and is not a stop node.

However, 
% considering the E(3)-invariance property of the property space and the equivariance property of the coordinates, 
we cannot apply the log-MSE loss directly. % for every iteration
Specifically, 
%if we take a E(3)-invariant representation perspective, 
for the first iteration $t=1$, there are only two nodes in the current subgraph, in which $\|\mathcal{V}_t\| = 2$ and $\mathcal{R}_t \in \mathbb{R}^{2\times 3}$. 
The only constraint to the current 3D structure is the distance between the two existing nodes. Thus, the ground truth coordinates can be considered as any point on a sphere with distance $d_t = \|\mathbf{r}_u - \bar{\mathbf{r}}_t\|$ to the reference point coordinate $\bar{\mathbf{r}}_t = \mathbf{r}_f$, 
where $\mathbf{r}_u$ denotes the ground truth coordinate of newly added node $u$ and $\mathbf{r}_f$ is the coordinate of the focus and the only node $f$ in previous subgraph. 
Moreover, for the situation that there are three nodes in the current subgraph, where $\|\mathcal{V}_t\| = 3$ and $\mathcal{R}_t \in \mathbb{R}^{3\times 3}$, the ground truth coordinate is considered equivalent to any point with a distance $d_t =\|\mathbf{r}_u - \bar{\mathbf{r}}_t\|$ to $\bar{\mathbf{r}}_t$ and a angle $\theta_t = \text{arccos}(\mathbf{s}_t, \mathbf{s}_{uf})$. Here, we use $\mathbf{s}_t, \mathbf{s}_{uf}$ to denote the unit vectors of $\mathbf{r}_u - \bar{\mathbf{r}}_t$ and $\mathbf{r}_u - \mathbf{r}_f$, respectively.
Therefore, if we use the log-MSE coordinate loss directly, all these situations will be neglected and the symmetric-invariance property of the geometric space will be violated.

To adapt E(3)-equivariant networks to \textit{de-novo} molecule generation task \textit{without external reference structure}, we propose to align the coordinates with Kabsch algorithm~\cite{kabsch1976solution} and then calculate the coordinate loss with transformed coordinates for the case when there are less than or equal to three nodes in any samples in this batch. Concretely, we calculate a rotation matrix $\mathbf{R} \in SO(3)$ and a translation vector $\tau \in \mathbb{R}^3$ for the maximum alignment between generated coordinates $\hat{\mathcal{R}}_t=\left[\hat{\mathbf{r}}_i\right]_{i\in \mathcal{V}_t}$ and ground truth coordinates $\mathcal{R}_t$. Then transformed coordinates of generated nodes are obtained by 
\begin{align}
    \tilde{\mathbf{r}}_i = \mathbf{R}\hat{\mathbf{r}}_i + \tau, \quad \forall i \in \mathcal{V}_t.
\end{align}
Thus, for samples with three or fewer nodes at the current iteration, the coordinate loss is:
% and the coordinate loss for iterations when there are less than or equal to three nodes in any samples in this batch is calculated as: 
\begin{align}
    \mathcal{L}_{\text{Coords}}^{t, i} = \sum_{j\in \mathcal{V}_{t,i}} \|\tilde{\mathbf{r}}_j - \mathbf{r}_j\|^2.
\end{align}
%  \frac{1}{|\mathcal{V}_t|} 
% So the total loss is the weighted sum of these individual losses.
% Formally,
% \begin{align}
%     \mathcal{L}_{\text{Total}} = \mathcal{L}_{\text{Prop}} + \alpha \mathcal{L}_{\text{Dis}} + \beta \mathcal{L}_{\text{Recon}} 
% \end{align}

\subsection{Generation with Explicit Control}
\label{sec: generation}
In the generation phase, we sample latent variables $\mathbf{z}_h$ and $\mathbf{z}_v$ and set the maximum number of fragments to $N$. Note that the \textit{exact} number of fragments might be smaller than $N$.
% During generation, we set a maximum number of fragments and sample same number of latent variables $\mathbf{z}_h$ and $\mathbf{z}_v$, although not all fragments might be included. 
Then the generation process follows the same decoding process introduced in Section~\ref{sec: decoder} but differs by operating without teacher forcing, relying solely on the model's self-guided predictions.
Notably, a unique feature of our E3WAE model is the disentangled latent space, which enables \textit{explicit control} over two key aspects of molecule generation:
(\romannumeral1) \textbf{property-targeting generation}: 
In this approach, pre-defined property latent variables $\mathbf{z}_{h,p}, \mathbf{z}_{v,p}$ are combined with sampled or template context latent variables $\mathbf{z}_{h,s}, \mathbf{z}_{v,s}$. 
This combination is then fed into the decoder, producing new molecules $\hat{\mathcal{G}}=\mathcal{D}(\mathcal{G}|\mathbf{z}_{h,p}, \mathbf{z}_{v,p})$ with targeted properties; 
(\romannumeral2) \textbf{context-preserving generation}: 
Conversely, this mode uses predefined context latent variables $\mathbf{z}_{h,s}, \mathbf{z}_{v,s}$ %are
combined with either sampled or template property latent variables $\mathbf{z}_{h,s}, \mathbf{z}_{v,s}$. 
Using such latent variables, the decoder can generate new molecules $\hat{\mathcal{G}}=\mathcal{D}(\mathcal{G}|\mathbf{z}_{h,s}, \mathbf{z}_{v,s})$ that refine certain properties while maintaining the core molecule framework.
The predefined latent variables are obtained either by directly using or by performing interpolation or extrapolation with latent variables from existing molecules. For simplicity, this work adopts the first approach.

\textbf{Atom Conformation Assembling.} Once the 3D molecular graph is complete at the fragment level, including all fragment types, links, and center coordinates, we proceed to determine the atom-level coordinates.
Following~\citet{qiang2023coarse}, we first randomly pick a fragment and explore all feasible links to its neighbors, selecting the one nearest to our pre-defined fragment center. We then use RDKit~\cite{landrum2016rdkit} to model each potential connection, evaluating their suitability based on root-mean-square deviation (RMSD) from the center. This procedure is iteratively applied to build the structure fragment by fragment. Finally, we align these local structures within the molecular framework using the Kabsch algorithm~\cite{kabsch1976solution}, adjusting the RDKit coordinates to match target positions.

\begin{table}[tbp]
\centering
\caption{Results on property-targeting generation. 
We tested the difference using mean squared error (MSE) and mean absolute error (MAE) between the given reference and the actual properties of the generated 3D molecules.
}\label{table:results_prop}
\vspace{-5 pt}
\resizebox{0.45\textwidth}{!}{
\begin{tabular}{llccccccccc}\toprule
\multicolumn{2}{c}{\multirow{2}{*}{}} &\multicolumn{2}{c}{Asphericity} &\multicolumn{2}{c}{QED} &\multicolumn{2}{c}{SAS} &\multicolumn{2}{c}{logP} \\\cmidrule{3-10}
& &MSE &MAE &MSE &MAE &MSE &MAE &MSE &MAE \\\midrule
\multirow{3}{*}{\shortstack[l]{GEOM}} 
&EDM &0.626 &0.455 &{0.113} &{0.285} &12.395 &3.385 &5.704 &1.903 \\
&HierDiff &{0.176} &{0.406} &0.12 &0.289 &{2.618} &{1.347} &\textbf{4.124} &\textbf{1.572} \\
&Ours &\textbf{0.095} &\textbf{0.246} &\textbf{0.072} &\textbf{0.221} &\textbf{1.563} &\textbf{1.002} &{4.490} &{1.630} \\
\midrule
\multirow{3}{*}{\shortstack[l]{Cross-\\Docked\\2020}} 
&EDM &0.109 &0.274 &0.147 &0.309 &11.244 &3.205 &{5.715} &1.957 \\
&HierDiff &{0.107} &{0.268} &{0.089} &{0.278} &{2.364} &{1.468} &5.752 &{1.897} \\
&Ours &\textbf{0.100} &\textbf{0.259} &\textbf{0.062} &\textbf{0.205} &\textbf{2.356} &\textbf{1.243} &\textbf{4.244} &\textbf{1.644} \\
\bottomrule
\end{tabular}}
\vspace{-10pt}
\end{table}

\section{Experiments}
\label{sec: exp}
In our work, we focus on generating large-scale drug-like molecules. We follow the experimental setup in \citet{qiang2023coarse} and conduct experiments mainly on two benchmark datasets for 3D molecule generation: GEOM-Drugs \cite{axelrod2022geom} and CrossDocked2020 \cite{francoeur2020three}. %Specifically, 
GEOM-Drugs contains 304k drug-like molecules while CrossDocked2020 consists of about 100k 3D ligand structures, each derived from a protein-ligand complex.
Our model is trained using a subset of 50,000 molecular structures from each dataset, specifically selecting those with the lowest energy conformations for each molecule.
We compare our model with three well-established baseline methods, including one auto-regressive method, G-SphereNet~\cite{luo2022an} and two equivariant diffusion-based methods, EDM~\cite{hoogeboom2022equivariant} and HierDiff~\cite{qiang2023coarse}.
Hyperparameters and other implementation details are provided in Appendix~\ref{appendix: additional_exp}. 

\vspace{-5pt}
\subsection{Property-targeting 3D Generation}
In AI-enhanced drug discovery, a crucial focus is the generation of 3D molecules with specific, desirable properties. This task is commonly referred to as \textit{property-targeting 3D generation}.
% There is a growing demand for inverse design on molecules. The goal is to find drug-like structures with desired properties, so-called property-targeting generation. 
Following the experiment settings in HierDiff~\cite{qiang2023coarse}, we select asphericity, QED, SAS, and logP as key properties for conducting practical drug discovery-related conditional generation.
% These properties enable the practical conditional generation of 3D molecules within the domain of drug discovery, with the detailed generation method introduced in Section~\ref{sec: generation}.
Asphericity assesses how much a molecule's shape deviates from a sphere, influencing its biological interactions. 
Quantitative Estimate of Drug-likeness (QED) gauges a molecule's similarity to known drugs, indicating its potential efficacy as a drug candidate. 
The Synthetic Accessibility Score (SAS) evaluates the ease of synthesizing a molecule, a crucial factor in drug development. 
Lastly, logP measures a compound's lipophilicity, which is essential for understanding the compound's absorption and distribution in the body. 
Each of these properties plays a significant role in determining a molecule's suitability for drug development, making them ideal choices for our focused properties.
% Each of these properties significantly influences a molecule's suitability for drug development, making them ideal choices for our model's focused generation.

For this task, as introduced in Section~\ref{sec: generation}, we combine the property latent embeddings of a reference molecule with another molecule's context latent embeddings to generate new molecules. 
To ensure a fair comparison and assess generalization, we use the properties of molecules in the test set as the target when generating new molecules for all baseline methods and our model.

\textbf{De-novo Molecule Generation.} 
We evaluate the mean squared error (MSE) and mean absolute error (MAE) between the input properties and the real properties of the generated 3D molecules. The results are shown in Table~\ref{table:results_prop}, and the detailed standard deviation of experiment results are provided in Appendix~\ref{appendix: std}.
The baseline results of Asphericity and QED on the GEOM-Drugs dataset are taken from HierDiff~
\cite{qiang2023coarse} while the others are reproduced by ourselves using their official implementation.
The results show that our model achieves the best performance on 7 out of 8 properties across two datasets. 
While we do not achieve the top ranking in the logP task, we attain results that closely approximate the current state-of-the-art method. 
The results demonstrate the effectiveness of our method in property-targeting generation for 3D drug-like molecules, showing that our model can serve as a useful tool in computational drug design.
We provide \textbf{results for general generation quality}, i.e., drug-likeness metrics and 3D conformation quality metrics, in Appendix~\ref{appendix: gen_qual}.
%, and \textbf{multiple property targeting generation} in Appendix \ref{appendix: multi-prop}.

\begin{table*}[tbp]
\centering
\caption{Results on property-guided ligand generation for target protein on CrossDocked2020 dataset. We reported the Vina Dock score for binding affinity and MSE and MAE between reference and actual properties of generated 3D molecules.}\label{table:results-bind}
\vspace{-10pt}
\resizebox{0.7\textwidth}{!}{
\begin{tabular}{lrrrrrrrrrrrrr}\toprule
&\multicolumn{3}{c}{Asphericity} &\multicolumn{3}{c}{QED} &\multicolumn{3}{c}{SAS} &\multicolumn{3}{c}{logP} \\\cmidrule{2-13}
&Vina &MSE &MAE &Vina &MSE &MAE &Vina &MSE &MAE &Vina &MSE &MAE \\\midrule
HierDiff &-4.253 &0.125 &0.296 &-4.429 &0.113 &0.275 &-5.053 &2.364 &1.620 &-4.293 &5.938 &1.855 \\
TargetDiff &-5.742 &0.117 &0.288 &-5.706 &0.112 &0.307 &-5.479 &3.369 &1.604 &-5.501 &4.509 &1.946 \\
Ours &\textbf{-5.891} &\textbf{0.102} &\textbf{0.271} &\textbf{-5.866} &\textbf{0.086} &\textbf{0.247} &\textbf{-5.940} &\textbf{2.358} &\textbf{1.529} &\textbf{-5.827} &\textbf{4.376} &\textbf{1.794} \\
\bottomrule
\end{tabular}}
\vspace{-10pt}
\end{table*}

\textbf{Structure-based Drug Design.} 
In real-world drug design applications, optimizing specific properties of binding molecules is a pivotal scenario. 
For instance, enhancing the synthetic accessibility score of a binding ligand can preserve its binding capability while simplifying the manufacturing process. 
Thus, we propose a property-targeting generation task for structure-based drug design. For this task, we train the model by taking pocket-ligand complex structures as input and performing ligand generation conditioned on a given property latent vector. Through this method, we generate 3D molecules while considering specific target binding sites and desired property values. 
We evaluate the generated molecules from targeted binding affinity and molecular properties. 
Following previous works% in this field
~\cite{guan2023d, luo20213d}, we employ AutoDock Vina~\cite{eberhardt2021autodock} to estimate the target binding affinity.
We use the MSE and MAE between generated and reference properties as property-related metrics. 
The results of are shown in Table~\ref{table:results-bind}. 
The results demonstrate that our model excels in generating binding ligands for target proteins with enhanced property values, surpassing the performance of the baseline model. This underscores the practical utility of our model in real-world structure-based drug design.

% It is always an important drug design scenario that we want to optimize certain properties eg. optimize the synthetic accessibility score of a binding ligand to a drug to maintain the binding property but make the drug easier to manufacture. 
% Thus we design this property-targeting generation for a structure-based drug design task.
% For this task, we take the pocket-ligand complex structure as input and conditional on a property latent vector.
% We perform 3D molecule generation conditional on given target binding sites and a property value. For this task, we evaluate one affinity-related metric, the Vina Dock score, and two property-related metrics, the MSE and MAE between generated and reference properties. The results are shown in Table~\ref{table:results-bind}.
% From the results, we can see our model is able to generate binding ligands to target proteins with improved property values.

\textbf{Multi-property Targeting Generation. }
Additionally, we apply our model to disentangle \textit{multiple} target properties and the structure context. Specifically, we conduct experiments with asphericity and QED properties using the GEOM-drug dataset. The results are shown in the following table. We can observe that the performance of multi-property targeting generation slightly dropped. Theoretically, asphericity and QED cannot be completely disentangled, but our model can still achieve a certain degree of explicit control on the multi-property-targeting generation.

\begin{table}[!htp]\centering
\vspace{-5pt}
\caption{Results on multi-property targeting generation with asphericity and QED properties on GEOM-drug. }\label{tab: multi-target}
\vspace{-10pt}
\resizebox{0.28\textwidth}{!}{
\begin{tabular}{lrrrrr}\toprule
&\multicolumn{2}{c}{Asphericity} &\multicolumn{2}{c}{QED} \\\cmidrule{2-5}
&MSE &MAE &MSE &MAE \\\midrule
EDM &0.594 &0.432 &0.147 &0.319 \\
HierDiff &0.193 &0.412 &0.136 &0.301 \\
E3WAE &\textbf{0.106} &\textbf{0.251} &\textbf{0.099} &\textbf{0.269} \\
\bottomrule
\vspace{-25pt}
\end{tabular}}
\end{table}

\begin{table}[bp]
\begin{center}
\vspace{-10 pt}
\caption{Results on context-preserving generation. The scores indicate the similarity of context latent variables (Emb.) and fingerprints (Fp.) between reference molecules and molecules generated with our model or retrieved from the training set. The top two results are highlighted as \textbf{1st} and \underline{2nd}.}
\vspace{-10pt}
\label{table:results_struct}
\resizebox{0.48\textwidth}{!}{
\begin{tabular}{llccccccccc}\toprule
\multicolumn{2}{c}{} &\multicolumn{2}{c}{Asphericity} &\multicolumn{2}{c}{QED} &\multicolumn{2}{c}{SAS} &\multicolumn{2}{c}{logP} \\\cmidrule{1-10}
& &Emb. &Fp. &Emb. &Fp. &Emb. &Fp. &Emb. &Fp. \\\midrule
\multirow{5}{*}{GEOM} &\textit{Ret. Mean} &0.33 &0.241 &0.403 &0.243 &0.472 &0.240 &0.437 &\underline{0.212} \\
&\textit{Ret. Max} &0.584 &\textbf{0.317} &\textbf{0.821} &\textbf{0.320} &\textbf{0.958} &\textbf{0.318} &\textbf{0.872} &\textbf{0.311} \\
\cmidrule{2-10}
&EDM &0.473 &0.130 &0.357 &0.135 &0.396 &0.126 &0.499 &0.131 \\
&HierDiff &0.419 &0.148 &0.391 &0.133 &0.514 &0.143 &0.463 &0.161 \\
&Ours &\textbf{0.593} &\underline{0.255} &\underline{0.618} &0.218 &\underline{0.532} &\underline{0.238} &\underline{0.624} &0.194 \\
\midrule
\multirow{5}{*}{\shortstack[l]{Cross-\\Docked\\2020}} &\textit{Ret. Mean} &0.260 &0.206 &0.331 &0.204 &\underline{0.496} &0.202 &0.503 &0.202 \\
&\textit{Ret. Max} &\underline{0.525} &\underline{0.280} &0.566 &\underline{0.271} &\textbf{0.974} &0.273 &\underline{0.726} &\underline{0.276} \\
\cmidrule{2-10}
&EDM &0.468 &0.152 &0.507 &0.140 &0.378 &0.131 &0.495 &0.147 \\
&HierDiff &0.461 &0.185 &\underline{0.587} &0.197 &0.412 &\underline{0.305} &0.484 &0.200 \\
&Ours &\textbf{0.570} &\textbf{0.344} &\textbf{0.665} &\textbf{0.307} &0.488 &\textbf{0.416} &\textbf{0.763} &\textbf{0.408} \\
\bottomrule
\end{tabular}}
\vspace{-15pt}
\end{center}
\end{table}

\vspace{-5pt}
\subsection{Context-Preserving 3D Generation}
\label{sec: context-preseving-results}
Our model %introduces
enables a novel \textit{structural context-preserving 3D generation} task, which optimizes targeted molecular properties while preserving the molecule's fundamental architecture. 
% This task provides advantages in lead optimization, where small adjustments to a molecule's structure can lead to significant improvements in its drug-like qualities. 
This capability plays a crucial role in enhancing the efficacy, safety, and synthesizability of molecules in drug development. 
For this task, as introduced in Section~\ref{sec: generation}, we fuse the context latent embeddings of a template molecule with another molecule's property latent embeddings to generate new molecules. 
The structural similarity between the template and the newly generated molecules is quantified by assessing the cosine similarity of their structural latent embeddings.
% We evaluate the cosine similarity between context latent embeddings of the template and the newly generated molecules as an indicator for structural similarity between molecules.

\textbf{Retrieval-based Baselines.} 
As \textit{no previous method can be adapted} to introduce a context condition and perform this context-preserving generation task, we design our baselines via retrieval and evaluate the metrics in the training set. 
Implementation details are in Appendix~\ref{appendix: retrieval}.
% Specifically, we first encode and save the context latent variables in the training set using our E3WAE structure encoder $\Theta_{s}$. Subsequently, we retrieve molecules that exhibit property values within a specified threshold. For comparison, we calculate and report both the mean (\textit{Ret. Mean}) and the maximum (\textit{Ret. Max}) similarity of structural embeddings between the template molecule and the molecules retrieved from the training set. This similarity metric serves as an indicator of structural resemblance, demonstrating the capability of our model in this task.

The embedding and fingerprint similarity scores between generated molecules and those obtained from retrieval-based baselines are presented in Table~\ref{table:results_struct}. 
% Note that the results of EDM and HierDiff are calculated by taking a molecule with same property values as the reference.
The scores for molecules generated by our model surpass nearly all of the mean retrieval similarity scores and, in some cases, even exceed the maximum retrieval values. 
This performance indicates that our model can generate new molecules that not only possess optimized properties but also closely resemble the reference template structure. 
Overall, this indicates the effectiveness of our model in context-preserving generation.
% The similarity scores of generated molecules and retrieval-based baselines are shown in Table~\ref{table:results_struct}. The scores of generated molecules outperform almost all retrieval mean and some even better than the retrieval max value, showing that our model is able generate molecules similar to the template molecule and optimized properties that are not in the training set.

% \todo{change to test set distributions}
\vspace{-5pt}
\subsection{Ablation Study}
We perform ablation studies to justify our model design on two key components of E3WAE: (1) model architecture and disentangled representation learning objectives, comparing WAE and VAE-based frameworks, and (2) the impact of using or omitting our proposed coordinate loss. The results demonstrate diminished performance or delayed convergence with alternatives to our model's design. Detailed results and analysis are in Appendix~\ref{appendix: ablation}.

% (1) the disentangle model architecture design: WAE \textit{vs.} VAE and (2) using \textit{vs.} without the proposed coordinate loss. Results display a decrease of performance or higher converging stage when using other design choices to our model. See Appendix~\ref{appendix: ablation} for details.
\vspace{-5pt}
\subsection{Visualization of Disentangled Latent Space}
% \vspace{-3pt}
% \begin{wrapfigure}[]{r}{0.52\textwidth}\vspace{-18pt}
% \begin{center}
% \includegraphics[width=0.52\textwidth]{figs/tsne/tsne.pdf}
% \vspace{-10pt}
% \caption{ t-SNE visualization of the model's disentangled latent spaces, colored by ground-truth property values.}
% \label{fig:tsne}
% \vspace{-15 pt}
% \end{center}
% \end{wrapfigure}
\begin{figure}[tbp]
\vspace{-5pt}
\begin{center}
\includegraphics[width=0.35\textwidth]{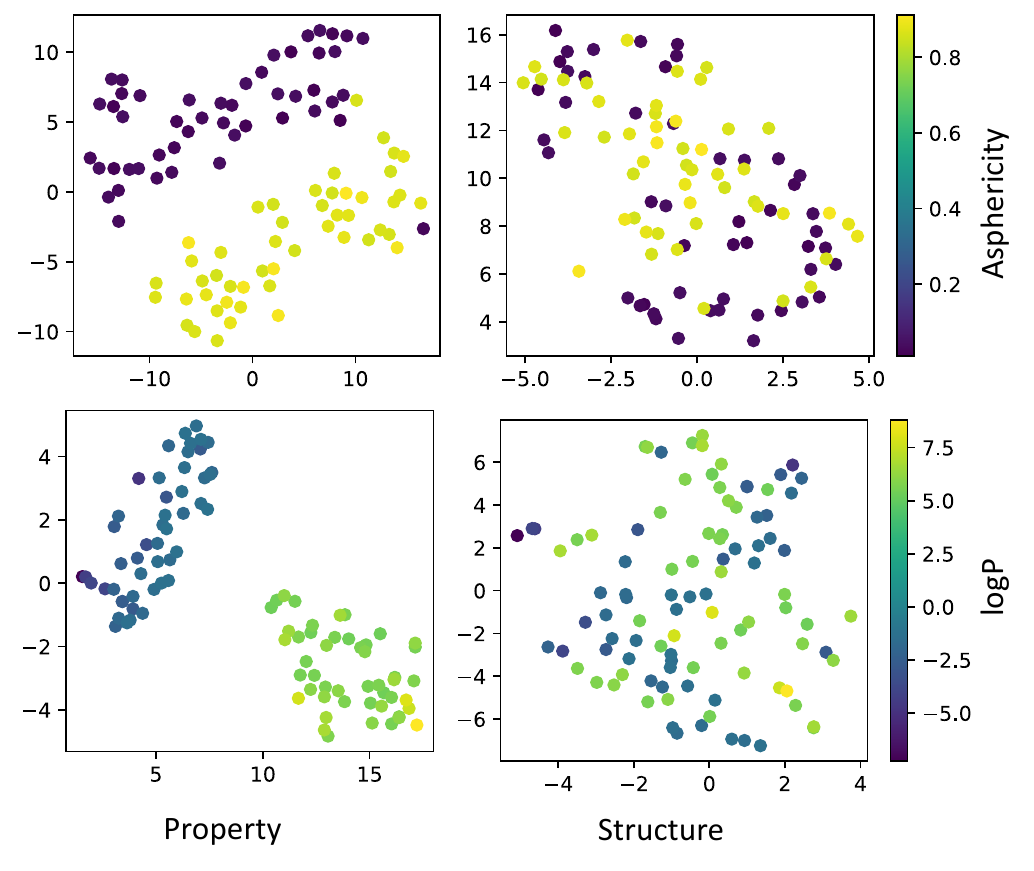}
\vspace{-10pt}
\caption{ t-SNE visualization of the model's disentangled latent spaces, colored by ground-truth property values.}
\label{fig:tsne}
\vspace{-20 pt}
\end{center}
\end{figure}
% To further demonstrate the disentangle ability of our model in a more clear way, we provide a t-SNE visualization~\cite{} of the latent variables, as shown in Figure~\ref{fig:tsne}.
To more explicitly show the disentangling ability of our model, we provide a t-SNE visualization~\cite{van2008visualizing} of the property and structure embeddings in Figure~\ref{fig:tsne}, focusing on asphericity and logP as examples. The color coding corresponds to various ground truth property values. 
% In the visualization, property latent variables form distinct clusters, whereas the context latent variables do not, indicating a mix. 
% This pattern reveals that our model effectively disentangles the property latent space in alignment with property labels, while the context latent space remains entangled with these labels. 
We observe that for each molecule, property labels can be well-represented by the property latent variables but not the context latent embeddings. The observed entanglement in the context latent space reflects the inherent independence of structural aspects from selected property labels, in line with the principles of disentangled representation learning.
This distinct separation in the property space demonstrates our model's ability to differentiate molecular properties and align them systematically, thereby enabling explicit control over generating 3D molecules with desired characteristics.
% The plots distinctly show separate clusters for property latent variables but mixed up for context latent variables.
% This indicates the well disentanglement of the property latent space according to property labels while context latent space does not. This make sense.
% This visual separation within the latent space confirms that our model not only distinguishes between different molecular characteristics but also organizes them in a coherent manner, facilitating the explicitly controlled generation of 3D molecules.

% Additional visualization of generated molecules are provided in Appendix~\ref{appendix: vis}.

\vspace{-5pt}
\section{Conclusion}
% We consider the conditional generation of 3D molecules with explicit control. 
% To achieve this, we propose an E(3)-equivariant Wasserstein autoencoder and factorize the latent space of a generative model into two disentangled aspects: the property and the structure context of 3D molecules.
% Our model ensures explicit control over these molecule attributes while maintaining equivariance of coordinates and invariance of likelihood. 
% Furthermore, to adapt equivariant networks for auto-regressive \textit{de-novo} 3D molecule generation from scratch, we introduce a novel alignment-based coordinate loss.
We present E3WAE, an E(3)-equivariant Wasserstein autoencoder for conditional generation of 3D molecules with explicit control. 
Extensive experiments on both over \textit{de-novo} 3D molecule generation and structure-based drug design validate our model's effectiveness in both property-guided and structure-guided molecule generation.
% Limitations and future work.
Additionally, our model can be adapted for explicit control over other drug design cases, such as % property/context-preserving generation for linker design %or
and protein sequence and structure co-design.
We leave these directions as future exploration.

%\newpage 

% \section*{Software and Data}
% Codes and data are included in the supplementary materials.

\section*{Acknowledgement}
We would like to thank Limei Wang for insightful discussions and feedback regarding the datasets used in this work.

\bibliography{camera-ready}

\begin{thebibliography}{61}
\providecommand{\natexlab}[1]{#1}

\bibitem[{Adams and Coley(2023)}]{adams2023equivariant}
Adams, K.; and Coley, C.~W. 2023.
\newblock Equivariant Shape-Conditioned Generation of 3D Molecules for Ligand-Based Drug Design.
\newblock In \emph{The Eleventh ICLR}.

\bibitem[{Axelrod and Gomez-Bombarelli(2022)}]{axelrod2022geom}
Axelrod, S.; and Gomez-Bombarelli, R. 2022.
\newblock GEOM, energy-annotated molecular conformations for property prediction and molecular generation.
\newblock \emph{Scientific Data}, 9(1): 185.

\bibitem[{Barber and Agakov(2004)}]{barber2004algorithm}
Barber, D.; and Agakov, F. 2004.
\newblock The im algorithm: a variational approach to information maximization.
\newblock \emph{NeurIPS}, 16(320): 201.

\bibitem[{Batatia et~al.(2022)Batatia, Kovacs, Simm, Ortner, and Cs{\'a}nyi}]{batatia2022mace}
Batatia, I.; Kovacs, D.~P.; Simm, G.; Ortner, C.; and Cs{\'a}nyi, G. 2022.
\newblock MACE: Higher order equivariant message passing neural networks for fast and accurate force fields.
\newblock \emph{NeurIPS}, 35: 11423--11436.

\bibitem[{Bengio et~al.(2021)Bengio, Jain, Korablyov, Precup, and Bengio}]{bengio2021flow}
Bengio, E.; Jain, M.; Korablyov, M.; Precup, D.; and Bengio, Y. 2021.
\newblock Flow network based generative models for non-iterative diverse candidate generation.
\newblock \emph{NeurIPS}, 34: 27381--27394.

\bibitem[{Bengio, Courville, and Vincent(2013)}]{bengio2013representation}
Bengio, Y.; Courville, A.; and Vincent, P. 2013.
\newblock Representation learning: A review and new perspectives.
\newblock \emph{IEEE transactions on pattern analysis and machine intelligence}, 35(8): 1798--1828.

\bibitem[{Brown et~al.(2019)Brown, Fiscato, Segler, and Vaucher}]{brown2019guacamol}
Brown, N.; Fiscato, M.; Segler, M.~H.; and Vaucher, A.~C. 2019.
\newblock GuacaMol: benchmarking models for de novo molecular design.
\newblock \emph{Journal of chemical information and modeling}, 59(3): 1096--1108.

\bibitem[{Burgess et~al.(2018)Burgess, Higgins, Pal, Matthey, Watters, Desjardins, and Lerchner}]{burgess2018understanding}
Burgess, C.~P.; Higgins, I.; Pal, A.; Matthey, L.; Watters, N.; Desjardins, G.; and Lerchner, A. 2018.
\newblock Understanding disentangling in $\beta$-VAE.
\newblock \emph{arXiv preprint arXiv:1804.03599}.

\bibitem[{Ceylan and Gutmann(2018)}]{ceylan2018conditional}
Ceylan, C.; and Gutmann, M.~U. 2018.
\newblock Conditional noise-contrastive estimation of unnormalised models.
\newblock In \emph{ICML}, 726--734. PMLR.

\bibitem[{Chen et~al.(2016)Chen, Duan, Houthooft, Schulman, Sutskever, and Abbeel}]{chen2016infogan}
Chen, X.; Duan, Y.; Houthooft, R.; Schulman, J.; Sutskever, I.; and Abbeel, P. 2016.
\newblock Infogan: Interpretable representation learning by information maximizing generative adversarial nets.
\newblock \emph{NeurIPS}, 29.

\bibitem[{Cheng et~al.(2020)Cheng, Min, Shen, Malon, Zhang, Li, and Carin}]{cheng2020improving}
Cheng, P.; Min, M.~R.; Shen, D.; Malon, C.; Zhang, Y.; Li, Y.; and Carin, L. 2020.
\newblock Improving Disentangled Text Representation Learning with Information-Theoretic Guidance.
\newblock In Jurafsky, D.; Chai, J.; Schluter, N.; and Tetreault, J., eds., \emph{Proceedings of the 58th Annual Meeting of the Association for Computational Linguistics}, 7530--7541. Online: Association for Computational Linguistics.

\bibitem[{Cover(1999)}]{cover1999elements}
Cover, T.~M. 1999.
\newblock \emph{Elements of information theory}.
\newblock John Wiley \& Sons.

\bibitem[{Deng et~al.(2021)Deng, Litany, Duan, Poulenard, Tagliasacchi, and Guibas}]{deng2021vector}
Deng, C.; Litany, O.; Duan, Y.; Poulenard, A.; Tagliasacchi, A.; and Guibas, L.~J. 2021.
\newblock Vector neurons: A general framework for so (3)-equivariant networks.
\newblock In \emph{ICCV}, 12200--12209.

\bibitem[{Deng et~al.(2020)Deng, Bakhtin, Ott, Szlam, and Ranzato}]{deng2020residual}
Deng, Y.; Bakhtin, A.; Ott, M.; Szlam, A.; and Ranzato, M. 2020.
\newblock Residual Energy-Based Models for Text Generation.
\newblock In \emph{ICLR}.

\bibitem[{Denton et~al.(2017)}]{denton2017unsupervised}
Denton, E.~L.; et~al. 2017.
\newblock Unsupervised learning of disentangled representations from video.
\newblock \emph{NeurIPS}, 30.

\bibitem[{Du et~al.(2022)Du, Guo, Wang, Shehu, and Zhao}]{du2022small}
Du, Y.; Guo, X.; Wang, Y.; Shehu, A.; and Zhao, L. 2022.
\newblock Small molecule generation via disentangled representation learning.
\newblock \emph{Bioinformatics}, 38(12): 3200--3208.

\bibitem[{Eberhardt et~al.(2021)Eberhardt, Santos-Martins, Tillack, and Forli}]{eberhardt2021autodock}
Eberhardt, J.; Santos-Martins, D.; Tillack, A.~F.; and Forli, S. 2021.
\newblock AutoDock Vina 1.2. 0: New docking methods, expanded force field, and python bindings.
\newblock \emph{Journal of chemical information and modeling}, 61(8): 3891--3898.

\bibitem[{Fey and Lenssen(2019)}]{Fey2019pyg}
Fey, M.; and Lenssen, J.~E. 2019.
\newblock Fast Graph Representation Learning with {PyTorch Geometric}.
\newblock In \emph{ICLR Workshop on Representation Learning on Graphs and Manifolds}.

\bibitem[{Francoeur et~al.(2020)Francoeur, Masuda, Sunseri, Jia, Iovanisci, Snyder, and Koes}]{francoeur2020three}
Francoeur, P.~G.; Masuda, T.; Sunseri, J.; Jia, A.; Iovanisci, R.~B.; Snyder, I.; and Koes, D.~R. 2020.
\newblock Three-dimensional convolutional neural networks and a cross-docked data set for structure-based drug design.
\newblock \emph{Journal of chemical information and modeling}, 60(9): 4200--4215.

\bibitem[{Fu et~al.(2021)Fu, Xiao, Li, Glass, and Sun}]{fu2021mimosa}
Fu, T.; Xiao, C.; Li, X.; Glass, L.~M.; and Sun, J. 2021.
\newblock Mimosa: Multi-constraint molecule sampling for molecule optimization.
\newblock In \emph{AAAI}, volume~35, 125--133.

\bibitem[{Garcia~Satorras et~al.(2021)Garcia~Satorras, Hoogeboom, Fuchs, Posner, and Welling}]{garcia2021n}
Garcia~Satorras, V.; Hoogeboom, E.; Fuchs, F.; Posner, I.; and Welling, M. 2021.
\newblock E (n) equivariant normalizing flows.
\newblock \emph{NeurIPS}, 34: 4181--4192.

\bibitem[{Gebauer, Gastegger, and Sch{\"u}tt(2019)}]{gebauer2019symmetry}
Gebauer, N.; Gastegger, M.; and Sch{\"u}tt, K. 2019.
\newblock Symmetry-adapted generation of 3d point sets for the targeted discovery of molecules.
\newblock \emph{NeurIPS}, 32.

\bibitem[{Geiger and Smidt(2022)}]{e3nn_paper}
Geiger, M.; and Smidt, T. 2022.
\newblock e3nn: Euclidean Neural Networks.

\bibitem[{Gretton et~al.(2012)Gretton, Borgwardt, Rasch, Sch{\"o}lkopf, and Smola}]{gretton2012kernel}
Gretton, A.; Borgwardt, K.~M.; Rasch, M.~J.; Sch{\"o}lkopf, B.; and Smola, A. 2012.
\newblock A kernel two-sample test.
\newblock \emph{The Journal of Machine Learning Research}, 13(1): 723--773.

\bibitem[{Guan et~al.(2023)Guan, Qian, Peng, Su, Peng, and Ma}]{guan2023d}
Guan, J.; Qian, W.~W.; Peng, X.; Su, Y.; Peng, J.; and Ma, J. 2023.
\newblock 3D Equivariant Diffusion for Target-Aware Molecule Generation and Affinity Prediction.
\newblock In \emph{The Eleventh ICLR}.

\bibitem[{Han et~al.(2021)Han, Min, Han, Li, and Zhang}]{han2021disentangled}
Han, J.; Min, M.~R.; Han, L.; Li, L.~E.; and Zhang, X. 2021.
\newblock Disentangled Recurrent Wasserstein Autoencoder.
\newblock In \emph{ICLR}.

\bibitem[{Higgins et~al.(2016)Higgins, Matthey, Pal, Burgess, Glorot, Botvinick, Mohamed, and Lerchner}]{higgins2016beta}
Higgins, I.; Matthey, L.; Pal, A.; Burgess, C.; Glorot, X.; Botvinick, M.; Mohamed, S.; and Lerchner, A. 2016.
\newblock beta-vae: Learning basic visual concepts with a constrained variational framework.
\newblock In \emph{ICLR}.

\bibitem[{Hoogeboom et~al.(2022)Hoogeboom, Satorras, Vignac, and Welling}]{hoogeboom2022equivariant}
Hoogeboom, E.; Satorras, V.~G.; Vignac, C.; and Welling, M. 2022.
\newblock Equivariant diffusion for molecule generation in 3d.
\newblock In \emph{ICML}. PMLR.

\bibitem[{Huang et~al.(2023)Huang, Sun, Du, and Lv}]{huang2023learning}
Huang, H.; Sun, L.; Du, B.; and Lv, W. 2023.
\newblock Learning Joint 2D \& 3D Diffusion Models for Complete Molecule Generation.
\newblock \emph{arXiv preprint arXiv:2305.12347}.

\bibitem[{Huang et~al.(2022)Huang, Peng, Ma, and Zhang}]{huang20223dlinker}
Huang, Y.; Peng, X.; Ma, J.; and Zhang, M. 2022.
\newblock 3DLinker: an E (3) equivariant variational autoencoder for molecular linker design.
\newblock \emph{arXiv preprint arXiv:2205.07309}.

\bibitem[{Igashov et~al.(2022)Igashov, St{\"a}rk, Vignac, Satorras, Frossard, Welling, Bronstein, and Correia}]{igashov2022equivariant}
Igashov, I.; St{\"a}rk, H.; Vignac, C.; Satorras, V.~G.; Frossard, P.; Welling, M.; Bronstein, M.; and Correia, B. 2022.
\newblock Equivariant 3d-conditional diffusion models for molecular linker design.
\newblock \emph{arXiv preprint arXiv:2210.05274}.

\bibitem[{Imrie et~al.(2020)Imrie, Bradley, van~der Schaar, and Deane}]{imrie2020deep}
Imrie, F.; Bradley, A.~R.; van~der Schaar, M.; and Deane, C.~M. 2020.
\newblock Deep generative models for 3D linker design.
\newblock \emph{Journal of chemical information and modeling}, 60(4): 1983--1995.

\bibitem[{Jin, Barzilay, and Jaakkola(2018)}]{jin2018junction}
Jin, W.; Barzilay, R.; and Jaakkola, T. 2018.
\newblock Junction Tree Variational Autoencoder for Molecular Graph Generation.
\newblock In \emph{ICML}, 2323--2332.

\bibitem[{Kabsch(1976)}]{kabsch1976solution}
Kabsch, W. 1976.
\newblock A solution for the best rotation to relate two sets of vectors.
\newblock \emph{Acta Crystallographica Section A: Crystal Physics, Diffraction, Theoretical and General Crystallography}, 32(5): 922--923.

\bibitem[{Kim and Mnih(2018)}]{kim2018disentangling}
Kim, H.; and Mnih, A. 2018.
\newblock Disentangling by factorising.
\newblock In \emph{ICML}, 2649--2658. PMLR.

\bibitem[{Kingma and Ba(2015)}]{kingma2014adam}
Kingma, D.~P.; and Ba, J. 2015.
\newblock Adam: A method for stochastic optimization.
\newblock In \emph{ICLR}.

\bibitem[{Kingma and Welling(2014)}]{kingma2013auto}
Kingma, D.~P.; and Welling, M. 2014.
\newblock {Auto-Encoding Variational Bayes}.
\newblock In \emph{2nd ICLR, {ICLR} 2014, Banff, AB, Canada, April 14-16, 2014, Conference Track Proceedings}.

\bibitem[{Kotovenko et~al.(2019)Kotovenko, Sanakoyeu, Lang, and Ommer}]{kotovenko2019content}
Kotovenko, D.; Sanakoyeu, A.; Lang, S.; and Ommer, B. 2019.
\newblock Content and style disentanglement for artistic style transfer.
\newblock In \emph{ICCV}, 4422--4431.

\bibitem[{Landrum(2016)}]{landrum2016rdkit}
Landrum, G. 2016.
\newblock RDKit: open-source cheminformatics http://www. rdkit. org.
\newblock 3(8).

\bibitem[{Lee et~al.(2018)Lee, Tseng, Huang, Singh, and Yang}]{lee2018diverse}
Lee, H.-Y.; Tseng, H.-Y.; Huang, J.-B.; Singh, M.; and Yang, M.-H. 2018.
\newblock Diverse image-to-image translation via disentangled representations.
\newblock In \emph{Proceedings of the European conference on computer vision (ECCV)}, 35--51.

\bibitem[{Li et~al.(2023)Li, Guo, Grazioli, Gerstein, and Min}]{li2023disentangled}
Li, T.; Guo, H.; Grazioli, F.; Gerstein, M.; and Min, M.~R. 2023.
\newblock Disentangled Wasserstein Autoencoder for T-Cell Receptor Engineering.
\newblock In \emph{Thirty-seventh Conference on Neural Information Processing Systems}.

\bibitem[{Liu et~al.(2022)Liu, Luo, Uchino, Maruhashi, and Ji}]{liu2022generating}
Liu, M.; Luo, Y.; Uchino, K.; Maruhashi, K.; and Ji, S. 2022.
\newblock Generating 3{D} Molecules for Target Protein Binding.
\newblock In Chaudhuri, K.; Jegelka, S.; Song, L.; Szepesvari, C.; Niu, G.; and Sabato, S., eds., \emph{Proceedings of the 39th ICML}, volume 162 of \emph{Proceedings of Machine Learning Research}, 13912--13924. PMLR.

\bibitem[{Locatello et~al.(2019)Locatello, Bauer, Lucic, Raetsch, Gelly, Sch{\"o}lkopf, and Bachem}]{locatello2019challenging}
Locatello, F.; Bauer, S.; Lucic, M.; Raetsch, G.; Gelly, S.; Sch{\"o}lkopf, B.; and Bachem, O. 2019.
\newblock Challenging common assumptions in the unsupervised learning of disentangled representations.
\newblock In \emph{ICML}, 4114--4124. PMLR.

\bibitem[{Luo et~al.(2021)Luo, Guan, Ma, and Peng}]{luo20213d}
Luo, S.; Guan, J.; Ma, J.; and Peng, J. 2021.
\newblock A 3D generative model for structure-based drug design.
\newblock \emph{NeurIPS}.

\bibitem[{Luo and Ji(2022)}]{luo2022an}
Luo, Y.; and Ji, S. 2022.
\newblock An Autoregressive Flow Model for 3D Molecular Geometry Generation from Scratch.
\newblock In \emph{ICLR}.

\bibitem[{Mollaysa, Paige, and Kalousis(2020)}]{mollaysa2020conditional}
Mollaysa, A.; Paige, B.; and Kalousis, A. 2020.
\newblock Conditional generation of molecules from disentangled representations.

\bibitem[{Paszke et~al.(2017)Paszke, Gross, Chintala, Chanan, Yang, DeVito, Lin, Desmaison, Antiga, and Lerer}]{paszke2017automatic}
Paszke, A.; Gross, S.; Chintala, S.; Chanan, G.; Yang, E.; DeVito, Z.; Lin, Z.; Desmaison, A.; Antiga, L.; and Lerer, A. 2017.
\newblock Automatic differentiation in pytorch.

\bibitem[{Polishchuk, G, and Varnek(2013)}]{polishchuk2013estimation}
Polishchuk, P.; G, T.~I., Madzhidov; and Varnek, A. 2013.
\newblock Estimation of the size of drug-like chemical space based on GDB-17 data.
\newblock \emph{Journal of computer-aided molecular design}, 27: 675--679.

\bibitem[{Powers et~al.(2022)Powers, Yu, Suriana, and Dror}]{powers2022fragmentbased}
Powers, A.~S.; Yu, H.~H.; Suriana, P.~A.; and Dror, R.~O. 2022.
\newblock Fragment-based ligand generation guided by geometric deep learning on protein-ligand structures.
\newblock In \emph{ICLR2022 Machine Learning for Drug Discovery}.

\bibitem[{Qiang et~al.(2023)Qiang, Song, Xu, Gong, Gao, Zhou, Ma, and Lan}]{qiang2023coarse}
Qiang, B.; Song, Y.; Xu, M.; Gong, J.; Gao, B.; Zhou, H.; Ma, W.-Y.; and Lan, Y. 2023.
\newblock Coarse-to-fine: a hierarchical diffusion model for molecule generation in 3d.
\newblock In \emph{ICML}, 28277--28299. PMLR.

\bibitem[{Satorras, Hoogeboom, and Welling(2021)}]{satorras2021n}
Satorras, V.~G.; Hoogeboom, E.; and Welling, M. 2021.
\newblock E (n) equivariant graph neural networks.
\newblock In \emph{ICML}, 9323--9332. PMLR.

\bibitem[{Shi et~al.(2019)Shi, Xu, Zhu, Zhang, Zhang, and Tang}]{shi2019graphaf}
Shi, C.; Xu, M.; Zhu, Z.; Zhang, W.; Zhang, M.; and Tang, J. 2019.
\newblock GraphAF: a Flow-based Autoregressive Model for Molecular Graph Generation.
\newblock In \emph{ICLR}.

\bibitem[{Sutherland et~al.(2017)Sutherland, Tung, Strathmann, De, Ramdas, Smola, and Gretton}]{sutherland2016generative}
Sutherland, D.~J.; Tung, H.-Y.; Strathmann, H.; De, S.; Ramdas, A.; Smola, A.; and Gretton, A. 2017.
\newblock Generative Models and Model Criticism via Optimized Maximum Mean Discrepancy.
\newblock In \emph{ICLR}.

\bibitem[{Tolstikhin et~al.(2018)Tolstikhin, Bousquet, Gelly, and Schoelkopf}]{tolstikhin2018wasserstein}
Tolstikhin, I.; Bousquet, O.; Gelly, S.; and Schoelkopf, B. 2018.
\newblock Wasserstein Auto-Encoders.
\newblock In \emph{ICLR}.

\bibitem[{Van~der Maaten and Hinton(2008)}]{van2008visualizing}
Van~der Maaten, L.; and Hinton, G. 2008.
\newblock Visualizing data using t-SNE.
\newblock \emph{Journal of machine learning research}, 9(11).

\bibitem[{Xie et~al.(2021)Xie, Shi, Zhou, Yang, Zhang, Yu, and Li}]{xie2021mars}
Xie, Y.; Shi, C.; Zhou, H.; Yang, Y.; Zhang, W.; Yu, Y.; and Li, L. 2021.
\newblock {\{}MARS{\}}: Markov Molecular Sampling for Multi-objective Drug Discovery.
\newblock In \emph{ICLR}.

\bibitem[{Xu et~al.(2023)Xu, Powers, Dror, Ermon, and Leskovec}]{xu2023geometric}
Xu, M.; Powers, A.~S.; Dror, R.~O.; Ermon, S.; and Leskovec, J. 2023.
\newblock Geometric latent diffusion models for 3d molecule generation.
\newblock In \emph{ICML}, 38592--38610. PMLR.

\bibitem[{Yang et~al.(2021)Yang, Hwang, Lee, Ryu, and Hwang}]{yang2021hit}
Yang, S.; Hwang, D.; Lee, S.; Ryu, S.; and Hwang, S.~J. 2021.
\newblock Hit and lead discovery with explorative rl and fragment-based molecule generation.
\newblock \emph{NeurIPS}, 34: 7924--7936.

\bibitem[{Yu(2020)}]{yu2020tutorial}
Yu, R. 2020.
\newblock A Tutorial on VAEs: From Bayes' Rule to Lossless Compression.
\newblock \emph{arXiv preprint arXiv:2006.10273}.

\bibitem[{Zhu et~al.(2018)Zhu, Zhang, Zhang, Wu, Torralba, Tenenbaum, and Freeman}]{zhu2018visual}
Zhu, J.-Y.; Zhang, Z.; Zhang, C.; Wu, J.; Torralba, A.; Tenenbaum, J.; and Freeman, B. 2018.
\newblock Visual object networks: Image generation with disentangled 3D representations.
\newblock \emph{NeurIPS}, 31.

\bibitem[{Zhu et~al.(2020)Zhu, Min, Kadav, and Graf}]{zhu2020s3vae}
Zhu, Y.; Min, M.~R.; Kadav, A.; and Graf, H.~P. 2020.
\newblock S3vae: Self-supervised sequential vae for representation disentanglement and data generation.
\newblock In \emph{CVPR}, 6538--6547.

\end{thebibliography}
\clearpage
\onecolumn
\section{Fragmentizing the Molecules}
\label{appendix: fragment}
To streamline the modeling of large molecules, fragment-based generation has been favored over atomic models, prompting the development of methods to segment molecules into fragments. An optimal decomposition algorithm should yield a comprehensive yet concise fragment vocabulary.

JT-VAE~\cite{jin2018junction} is the pioneer work for fragment-based molecule generation, using a minimum spanning tree algorithm to decompose molecules while maintain chemical bonds and avoid cycles. It achieved coverage of all synthetically accessible structures with a vocabulary under 800. Subsequent models like MARS~\cite{xie2021mars}, FREED~\cite{yang2021hit}, and MIMOSA~\cite{fu2021mimosa}, each develop unique bond-breaking criteria, and further refines the vocabulary while ensuring frequency-based relevance. 
A recent study, HierDiff~\cite{qiang2023coarse}, expands upon this approach by leveraging the tree decomposition algorithm from JT-VAE for fragment-based 3D molecule generation, introducing center coordinates as the three-dimensional representation for fragments.

Given the vast chemical space, it is practical to exclude less common fragments. % omitting rare fragments is pragmatic. 
Following HierDiff~\cite{qiang2023coarse}, we employ \citet{jin2018junction} tree decomposition in 3D space, which includes the following steps: 
% (1) extracting  the set of chemical bonds which do not belongs to any rings and identifying the set of simple rings which only represent a single topological cycle from the molecules; 
% (2) merging bridged rings that share more than two atoms as the bridged rings are chemically significant and exhibit unique 3D shapes; 
% (3) adding the intersecting atom which connects more than 3 bonds to the graph as a fragment, preventing the formation of cycles; and finally, 
% (4) assigning 3D geometric representation based on the coordinates of each fragment's geometric center. 
Firstly, we extract chemical bonds not forming part of any rings and identify isolated rings representing single topological cycles. Secondly, we merge bridged rings that share over two atoms, acknowledging their chemical importance and distinct 3D conformations. Thirdly, to avoid cyclic structures, any intersecting atom bonded to more than three others is incorporated into the graph as an individual fragment. Finally, we provide each fragment with a 3D geometric representation based on the centroid of its atoms.
By this fragmentizing method, we balance structural complexity with the need for a reasonable-sized fragment vocabulary. 
For the encoder, we use the high-level features based on property values as outlined in HierDiff, along with the embeddings of fragment types, as the input features for each fragment.

% 1. Extraction of chemical bonds that do not belong to any rings and identification of simple rings representing single topological cycles within the molecules.
% 2. The merging of bridged rings that share more than two atoms, as bridged rings hold significant chemical relevance and display distinct 3D shapes.
% 3. Addition of intersecting atoms that connect more than three bonds to the graph as fragments, preventing the formation of cycles.
% 4. Assignment of 3D geometric representations based on the coordinates of each fragment's geometric center.

\section{Disentanglement Guarantee}
\label{appendix: disentanglement_guarantee}

In this section, we provide a theoretical analysis to demonstrate our model's capacity for guaranteeing the disentanglement of property and context latent variables.  
Following previous work on disentangled representation learning~\cite{cheng2020improving, han2021disentangled, li2023disentangled}, we define a measurement of embedding disentanglement as follows, 
\begin{align}
    \text{D}(\mathbf{z}_p, \mathbf{z}_s; \mathbf{x}) = \text{VI}(\mathbf{z}_p; \mathbf{x}) + \text{VI}(\mathbf{z}_s; \mathbf{x}) - \text{VI}(\mathbf{z}_p; \mathbf{z}_s),
\end{align}
where $\mathbf{z}_p$ and $\mathbf{z}_s$ are the property and context latent variables, respectively, and $\mathbf{x}$ is the input molecule.
Variation of Information, $\text{VI}$, is a measure of independence between two random variables.
The mathematical definition of VI between random variables $\boldsymbol{x}$ and $\boldsymbol{y}$ is
\begin{align}
    \text{VI}(\boldsymbol{x}; \boldsymbol{y}) = \text{H}(\boldsymbol{x}) + \text{H}(\boldsymbol{y}) - 2\text{I}(\boldsymbol{x}; \boldsymbol{y}),
\end{align}
where $\text{H}(\boldsymbol{x})$ and $\text{H}(\boldsymbol{y})$ are entropies of $\boldsymbol{x}$ and $\boldsymbol{y}$, respectively, and
$\text{I}(\boldsymbol{x}; \boldsymbol{y})$ is the mutual information (MI) between variables $\boldsymbol{x}$ and $\boldsymbol{y}$.
Additionally, $\text{VI}(\boldsymbol{x}; \boldsymbol{y}) = 0$ indicates $\boldsymbol{x}$ and $\boldsymbol{y}$ are the sample variable.
Intuitively, the mutual information is a measure of ``dependence'' between two variables, 
and the VI distance is a measure of ``independence'' between them.

This measurement reaches 0 when $\mathbf{z}_p$ and $\mathbf{z}_s$ are totally independent, \textit{i.e.}, fully disentangled. Thus, we treat this measurement as our disentangling objective.

By the definition of VI, the measurement $D$ can be simplified as:
\begin{align}
    \text{VI}(\mathbf{z}_p; \mathbf{x}) + \text{VI}(\mathbf{z}_s; \mathbf{x}) - \text{VI}(\mathbf{z}_p; \mathbf{z}_s) 
    = 2\text{H}(\mathbf{x}) 
    + 2\left[
    \text{I}(\mathbf{z}_p; \mathbf{z}_s) - \text{I}(\mathbf{z}_p; \mathbf{x}) - \text{I}(\mathbf{z}_s; \mathbf{x}) 
    \right].
\end{align}
Since $\text{H}(\mathbf{x})$ is a constant associated with the data,
we only need to focus on $\text{I}(\mathbf{z}_p; \mathbf{z}_s) - \text{I}(\mathbf{z}_p; \mathbf{x}) - \text{I}(\mathbf{z}_s; \mathbf{x}) $.

Giving rise to the problem that without any inductive bias in supervision,
the disentangled representation could be meaningless~\cite{locatello2019challenging}, we add inductive biases by using the property value $y$ as supervised information for the property latent $\mathbf{z}_p$. 
As $\mathbf{z}_p \rightarrow \mathbf{x} \rightarrow y$ forms a Markov chain, we have $\text{I}(\mathbf{z}_p; \mathbf{x}) \geq \text{I}(\mathbf{z}_p; y)$ based on the MI data-processing inequality~\cite{cover1999elements}.
Therefore, we derive a upper bound for the measurement as:
\begin{align}
    \text{I}(\mathbf{z}_p; \mathbf{z}_s) - \text{I}(\mathbf{z}_p; \mathbf{x}) - \text{I}(\mathbf{z}_s; \mathbf{x}) 
    \leq \text{I}(\mathbf{z}_p; \mathbf{z}_s) - \text{I}(\mathbf{z}_p; y) - \text{I}(\mathbf{z}_s; \mathbf{x}).
    \label{eq: upper_bound}
\end{align}
Next, we show how our model design and training objective aligns to the optimization of the terms $\text{I}(\mathbf{z}_p; \mathbf{z}_s)$,  $\text{I}(\mathbf{z}_p; y)$, and $\text{I}(\mathbf{z}_s; \mathbf{x})$ in Equation~\ref{eq: upper_bound}.

\paragraph{Minimizing $\text{I}(\mathbf{z}_p; \mathbf{z}_s)$. }
By minimizing the Wasserstein regularization term $\mathcal{L}_{\text{Dis}}$, we forces the distribution of the combined latent variables $\mathbf{z} = \text{concat}(\mathbf{z}_p, \mathbf{z}_s)$ to approach an isotropic Gaussian distribution, where all
dimensions are independent. This enforces the independence across all dimensions of $\mathbf{Z}$. Consequently, it minimizes the mutual information between the latent variables $\mathbf{z}_p$ and $\mathbf{z}_s$, aiding in their disentanglement.

\paragraph{Maximizing $\text{I}(\mathbf{z}_p; y)$. } By leveraging an MI variational lower bound~\cite{barber2004algorithm}, we derive a lower bound for $\text{I}(\mathbf{z}_p; y)$ as
\begin{align}
    \text{I}(\mathbf{z}_p; y) \geq \text{H}(y) + \mathbb{E}_{\pi(y, \mathbf{z}_p)}\left[ \log q_{\mathcal{H}}(y|\mathbf{z}_p) \right].
\end{align}
Concretely, the latent variable $\mathbf{z}_p$ is encouraged to give a better prediction of property value $y$ by maximizing $\mathbb{E}_{\pi(y, \mathbf{z}_p)}\left[ \log q_{\mathcal{H}}(y|\mathbf{z}_p) \right]$, where $q_{\mathcal{H}}(y|\mathbf{z}_p)$ is the predicted probability by the auxiliary prediction head $\mathcal{H}$. 
Thus, optimizing the performance of $\mathcal{H}$ maximizes $\text{I}(\mathbf{z}_p; y)$.

\paragraph{Maximizing $\text{I}(\mathbf{z}_s; \mathbf{x})$. }
According to Theorem 3 in \citet{han2021disentangled} and Theorem 2.1 in \citet{li2023disentangled}, given the encoder $Q_\theta(\mathbf{z}|\mathbf{x})$, prior $P_{\mathbf{z}}$ and the data distribution $P_D$, we have
\begin{align}
    \mathbb{D}_{\text{KL}}(Q_{\mathbf{z}}\| P_{\mathbf{z}}) 
    = \mathbb{E}_{P_D} \left[ \mathbb{D}_{\text{KL}}\left( Q_\theta(\mathbf{z}|\mathbf{x}) \| P_{\mathbf{z}} \right) 
    \right] - \text{I}(\mathbf{z}; \mathbf{x}),
    \label{eq: theorem}
\end{align}
where $\text{KL}$ represents the Kullback-Leibler divergence and the $Q_{\mathbf{z}}$ is the marginal distribution of the encoder when $\mathbf{x} \sim P_D$ and $\mathbf{z} \sim Q_\theta(\mathbf{z}|\mathbf{x})$.
Equation~\ref{eq: theorem} demonstrate that by minimizing the KL divergence between $Q_{\mathbf{z}}$ and $P_{\mathbf{z}}$, we jointly maximize the mutual information between the data $\mathbf{x}$ and the latent variable $\mathbf{z}$.
This also applies to the two separate parts of $\mathbf{z}$, $\mathbf{z}_p$ and $\mathbf{z}_s$. 
In practice, measuring the marginal directly is often infeasible. Consequently, we resort to minimizing the kernel MMD outlined in Equation~\ref{eq: mmd}. 
This approach obviates the necessity for additional constraints on the information content of $\mathbf{z}_s$, as $\text{I}(\mathbf{z}_s; \mathbf{x})$ is inherently maximized by the objective.

\section{3D Molecule Graph Reconstruction Algorithm}
\label{appendix: decode_alg}

\begin{figure}[h]
    \centering
    \includegraphics[width=0.6\textwidth]{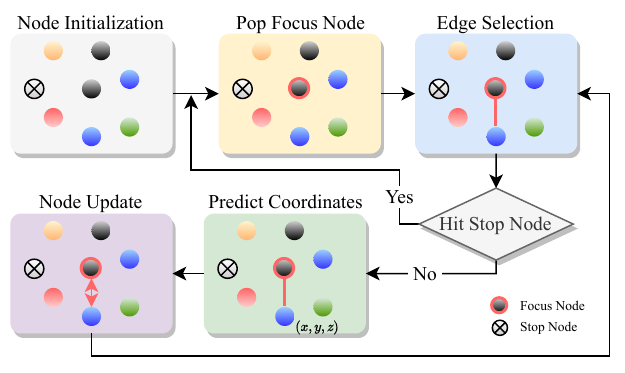}
    \caption{An illustration of the reconstruction process. }
    \label{fig: recon}
\end{figure}

The 3D molecule graph reconstruction process, i.e., the decoding process, is illustrated in Figure~\ref{fig: recon} Algorithm~\ref{alg:reconstruct}.

\algnewcommand{\Initialize}[1]{%
	\State \textbf{Initialize:}
	\Statex \hspace*{\algorithmicindent}\parbox[t]{.8\linewidth}{\raggedright #1}
}
\algnewcommand\algorithmicand{\textbf{and}}
\algnewcommand\algorithmicnot{\textbf{not}}
\algnewcommand{\algvar}[1]{{\text{\ttfamily\detokenize{#1}}}}
\algnewcommand\RETURN{\State \algorithmicreturn}%

\begin{algorithm}[h]
\caption{3D Molecule Graph Reconstruction in E3WAE}
\label{alg:reconstruct}
\begin{algorithmic}[1]
\State\textbf{Input:} Latent variables $\mathbf{z}_{h}$, $\mathbf{z}_{v}$
% \Ensure Reconstructed 3D molecular graph $\mathcal{G}$

\Initialize{
$\{x_i\}_{i=1,\cdots, n} \leftarrow \text{NodeTypes}(\mathbf{z}_{h}$, $\mathbf{z}_{v})$ \\
Queue $Q \leftarrow \emptyset$ \\
$Q.\text{push}(\text{RandomSelect}(\{1,\cdots, n\})$ \\
% \quad~     \Comment{\small  Initialize the focus queue by randomly select a node}\\
3D Graph $\mathcal{G} = \{\mathcal{V}, \mathcal{E}, \mathcal{R}\}$, where
$\mathcal{V} \leftarrow \emptyset$, 
$\mathcal{E} \leftarrow \emptyset$, 
$\mathcal{R} \leftarrow \emptyset$ 
}

\While{$Q \neq \emptyset$}
    \State $f \leftarrow Q.\text{pop}()$
    \State Add node $f$ to graph $\mathcal{G}$
    \State $\text{isStopNode} \leftarrow \text{False}$
    \While{\algorithmicnot $\text{ isStopNode}$}
        \State $(i, \text{isStopNode}) \leftarrow \text{PredictEdge}(f, \mathcal{G})$
        \If{\algorithmicnot $\text{ isStopNode}$ \algorithmicand $\text{ FirstLink}(i)$}
        \State $\mathbf{r}_u = \left(x_u, y_u, z_u\right) \leftarrow \text{PredictCoords}(i)$
        \State $\mathcal{V} \leftarrow \mathcal{V} \cup \{i\}$, $\mathcal{E} \leftarrow \mathcal{E} \cup \{(f, i)\}$, 
        \State $\mathcal{R} \leftarrow \mathcal{R} \cup \{\mathbf{r}_i\}$
        \State $Q.\text{push}(i)$
        \EndIf
    \State $\text{MarkVisited}(f)$
    \EndWhile
\EndWhile
\State\textbf{Return: } Reconstructed $\mathcal{G}= (\mathcal{V}, \mathcal{E}, \mathcal{R})$

\end{algorithmic}
\end{algorithm}

% \paragraph{Node Type Prediction }
% % \paragraph{$\text{NodeTypes}(\mathbf{z}_{h}$, $\mathbf{z}_{v})$}
% The fragment type logits are obtained from latent variables through a self-attention mechanism and an MLP, and these logits are then used to sample the fragment types. Once these types are determined, their embeddings are concatenated with the corresponding invariant latent variables $\mathbf{z_h}$ for further steps. Here, we adopt a cross-entropy loss $\mathcal{L}_{\text{NodeType}}$ to optimize fragment type prediction.

% \paragraph{Edge Prediction}
% % \paragraph{$\text{PredictEdge}(v, \mathcal{G})$}
% The node focusing and edge connecting order is pre-determined by a Breath-First
% Search to enable teacher-forcing training. For each focusing node 

% \paragraph{Coordinates Prediction}
% % \paragraph{$\text{PredictCoords}(u)$}
% For each newly added node to the graph, we predict its coordinates using the 

\section{Dataset Description}
\label{appendix: dataset}
\paragraph{GEOM-Drugs} GEOM-Drugs~\cite{axelrod2022geom} is a large-scale dataset that contains molecules with up to 181 atoms and an average of 44.4 atoms. 
GEOM-Drugs provides multiple conformations for each molecule with corresponding energies, and we retain one stable conformation with the lowest energy to construct the dataset. 
We take the first 50k molecules to train our model.
The training/validation/test split ratio is 70\%, 15\%, and 15\%, respectively.

\paragraph{CrossDocked2020} 
CrossDocked2020~\cite{francoeur2020three} is a dataset designed for the development and benchmarking of structure-based drug discovery. It comprises 22.5 million poses of ligands docked into a variety of similar binding pockets sourced from the Protein Data Bank. 
Our study focus on the ligand structures and corresponding drug-like properties derived from this dataset.
For model training, we select a subset of 50k molecules.
The training/validation/test split ratio is 70\%, 15\%, and 15\%, respectively.

\section{Additional Experimental Details}
\label{appendix: additional_exp}
This section describes the full experiment setup in this paper. The implementation of our methods is based on the PyTorch~\cite{paszke2017automatic} and Pytorch Geometric~\cite{Fey2019pyg}, and all models are trained with the Adam optimizer~\cite{kingma2014adam}. 
All experiments are conducted on a single NVIDIA A100 80GB GPU. 
The search space for model and training hyperparameters are listed in Table~\ref{tab: hyperpara}. Note that we select hyperparameters for experiments on asphericity, QED, SA Score, logP properties by the same search space, and the optimal hyperparameters are chosen by the performance on the validation set.

\begin{table}[ht]\centering
\caption{Model and training hyperparameters for our method on different datasets.}\label{tab: hyperpara}
\resizebox{0.7\textwidth}{!}{
\begin{tabular}{lccccc}\toprule
\multirow{2}{*}{\textbf{Hyperparameter}} &\multicolumn{2}{c}{\textbf{Values/Search Space}} \\
\cmidrule{2-3}
&\textbf{GROM-Drugs} &\textbf{CrossDocked2020} \\\midrule
Number of layers &4, 5, 6, 7, 8, 9 &4, 6, 9 \\
Hidden dim &28, 32 &28, 32 \\
Coordinate loss trade-off weight &0.5, 1, 3 &0.5, 1, 3 \\
Property prediction loss trade-off weight &3, 5, 10, 15 &5, 10, 15 \\
Wasserstein loss trade-off weight &3, 5, 10 &5, 10, 15 \\
Weight Decay &1e-4 &1e-4 \\
% Dropout &0.2, 0.3, 0.5 &0.2, 0.3, 0.5 \\
Epochs &50 &50 \\
Batch size &16, 32 &16, 32 \\
Learning rate &1e-4, 2e-4, 5e-4, 1e-3 &1e-4, 2e-4, 5e-4 \\
Learning rate decay factor &0.1, 0.5 &0.1, 0.5 \\
Learning rate decay epochs &3, 5, 10 &3, 5, 10, 20 \\
\bottomrule
\end{tabular}}
\end{table}

\subsection{Implementation of Baselines} \label{appendix: reproduce_baselines}
\paragraph{EDM.}
For EDM~\cite{hoogeboom2022equivariant}, we use the official implementation available on GitHub\footnote{\url{https://github.com/ehoogeboom/e3_diffusion_for_molecules}, The MIT License.}. 
We adhered to the default hyperparameters specified in the repository. 
To facilitate property-targeting generation, we train EDM models to incorporate drug-like property values as an additional input, concatenated to the node features.
During the generation phase, we sample the number of nodes from the node distribution of the training set and employ property values from the test set as reference for generating new 3D molecules.

\paragraph{HierDiff.}
We implement HierDiff~\cite{qiang2023coarse} using its official GitHub repository\footnote{\url{https://github.com/qiangbo1222/HierDiff}, The MIT License.}, following the prescribed default hyperparameters. 
For property-targeting generation, we trained both the coarse-grained diffusion and edge denoising generation models, integrating drug-like property values into the node features as context. 
During generation, the fragment numbers was determined based on the distribution observed in the training set, with property values from the test set guiding the generation of 3D molecules.

\paragraph{TargetDiff.}
We implement TargetDiff~\cite{guan2023d} with its official GitHub repository\footnote{\url{https://github.com/guanjq/targetdiff}, The MIT License.}, following the prescribed default hyperparameters. 
For adapting the original work to property-targeting generation, we use a straightforward extension similar to conditional EDM~\cite{hoogeboom2022equivariant} Section 3.4 and Appendix E. 
Specifically, a conditioning on a property value $c$ is added where relevant. 
The diffusion process that adds noise is not altered. 
The generative denoising process is conditioned on $c$ by adding it as input to the neural network.

Note that, for context-preserving generation, the results of EDM and HierDiff in Section~\ref{sec: context-preseving-results} are calculated by taking a molecule with same property values as the reference, since they cannot take the template structure as guidance for conditional generation.

\subsection{Implementation of Retrieval-based Baselines}
\label{appendix: retrieval}
Specifically, we first encode and save the context latent variables in the training set using our E3WAE structure encoder $\Theta_{s}$. Subsequently, we retrieve molecules that exhibit property values within a specified threshold. For comparison, we calculate and report both the mean (\textit{Ret. Mean}) and the maximum (\textit{Ret. Max}) similarity of structural embeddings between the template molecule and the molecules retrieved from the training set. This similarity metric serves as an indicator of structural resemblance, demonstrating the capability of our model in this task.

\subsection{Standard deviation on property-targeting generation}
\label{appendix: std}
To quantify the variability and reliability of our results, we present the standard deviation of results for property-targeting generation in Table~\ref{table:results_prop_std}.

\begin{table*}[ht]
\centering
\caption{Standard deviation on property-targeting generation. 
}\label{table:results_prop_std}
\vspace{-5 pt}
\resizebox{0.65\textwidth}{!}{
\begin{tabular}{llccccccccc}\toprule
\multicolumn{2}{c}{\multirow{2}{*}{}} &\multicolumn{2}{c}{Asphericity} &\multicolumn{2}{c}{QED} &\multicolumn{2}{c}{SAS} &\multicolumn{2}{c}{logP} \\\cmidrule{3-10}
& &MSE &MAE &MSE &MAE &MSE &MAE &MSE &MAE \\\midrule
\multirow{3}{*}{GEOM-Drugs} &EDM &0.088 &0.153 &0.087 &0.153 &6.317 &0.965 &7.427 &1.442 \\
&HierDiff &0.096 &0.166 &0.124 &0.186 &1.908 &0.759 &2.531 &0.855 \\
&Ours &0.075 &0.130 &0.071 &0.134 &2.215 &0.747 &4.900 &1.241 \\
\midrule
\multirow{3}{*}{CrossDocked2020} &EDM &0.083 &0.150 &0.131 &0.179 &6.368 &0.983 &7.199 &1.372 \\
&HierDiff &0.082 &0.147 &0.069 &0.132 &1.567 &0.645 &8.348 &1.468 \\
&Ours &0.095 &0.123 &0.073 &0.142 &1.775 &0.701 &6.008 &1.241 \\
\bottomrule
\end{tabular}}
\vspace{-5pt}
\end{table*}

\subsection{General Generation Quality}\label{appendix: gen_qual}
In this section, we evaluate and report the general generation quality of our model, focusing on two key aspects: drug-likeness and 3D conformation quality.

\textbf{Drug-Likeness Evaluation.}
The drug-likeness of generated molecules is a critical measure of their potential as drug candidates. 
Following the experiment settings in \citet{qiang2023coarse}, we evaluate our model using additional metrics: Retrosynthetic Accessibility (RA), Medicinal Chemistry Filter (MCF)~\cite{brown2019guacamol}, $\Delta$logP, and $\Delta$MW. RA assesses the synthetic feasibility of molecules, while MCF quantifies the percentage of molecules devoid of undruggable substructures. $\Delta$logP measures the deviation of logP values from the ground truth of the training set, and $\Delta$MW evaluates the variance in Molecular Weight from the training set's standard.
Our model shows competitive performance in these metrics, particularly excelling in MCF scores and showing close proximity to SOTA methods in others.
Moreover, it's important to mention that these baseline methods typically require the input of a node number derived from the node distribution, a condition that is often impractical in real-world scenarios. 
These results demonstrate the model's capability to effectively generate drug-like molecules.
\begin{table}[h]
% \vspace{-10 pt}
\begin{center}
% \vspace{-5 pt}
\caption{Results on drug-likeness evaluation,
% The top two results are highlighted as \textbf{1st} and {2nd}. 
where the last column stands for the results of training data. 
* denotes methods that contain a diffusion model and need to take a node number from the node distribution as input. 
GSNet is short for G-SphereNet. 
Experiments that report errors due to low validity
are marked as ’–’.}
\label{table:results_druglike}
% \vspace{-5 pt}
\resizebox{0.5\textwidth}{!}{
\begin{tabular}{lcccccc}\toprule
&GSNet &EDM* &HierDiff* &Ours &GEOM-Drugs \\\midrule
QED $\uparrow$ &0.382 &0.505 &{0.509} &\textbf{0.527} &0.658 \\
RA $\uparrow$ &-{}- &0.368 &{0.611} &\textbf{0.559} &0.915 \\
MCF $\uparrow$ &0.489 &0.552 &{0.654} &\textbf{0.763} &0.774 \\
SAS $\downarrow$ &-{}- &5.346 &{4.500} &\textbf{4.077} &4.018 \\
$\Delta$logP $\downarrow$ &2.306 &0.577 &{0.279} &\textbf{0.194} &0.000 \\
$\Delta$MW $\downarrow$ &170.700 &{24.578} &\textbf{19.781} &69.085 &0.000 \\
\midrule
&GSNet &EDM* &HierDiff* &Ours &CrossDocked \\
\midrule
QED $\uparrow$ &0.442 &0.402 &\textbf{0.486} &{0.435} &0.619 \\
RA $\uparrow$ &-{}- &0.358 &{0.585} &\textbf{0.716} &0.912 \\
MCF $\uparrow$ &0.449 &0.416 &\textbf{0.726} &{0.645} &0.746 \\
SAS $\downarrow$ &-{}- &7.325 &{4.536} &\textbf{4.328} &2.564 \\
$\Delta$logP $\downarrow$ &3.359 &1.576 &{1.132} &\textbf{0.851} &0.000 \\
$\Delta$MW $\downarrow$ &200.950 &{28.856} &\textbf{4.763} &76.801 &0.000 \\
\bottomrule
\end{tabular}}
\end{center}
% \vspace{-15pt}
\end{table}

\textbf{Conformation Quality Evaluation.}
The evaluation of 3D generation quality is conducted using stability metrics and substructure geometry alignment metrics. Stability metrics, as introduced by \citet{hoogeboom2022equivariant}, includes atomic stability (the proportion of atoms with correct valency) and molecular stability (the proportion of molecules where all atoms are stable). For the substructure geometry alignment metrics, following~\citet{huang2023learning}, we select the 8 most frequent types of bonds, bond pairs, and bond triples, and compute the MMD distances on bond lengths, bond angles, and dihedral angles distributions between generated molecules and test set. 
As shown in Table~\ref{table:results_3d}, the performance of our model either surpass or are on par with those of diffusion-based methods, known for their high computational complexity. This indicates the model's effective generation of molecules with not only realistic but also structurally stable 3D conformations.
% The 3D generation quality, as shown in Table~\ref{table:results_3d}, is evaluated based on stability metrics as introduced in~\citet{hoogeboom2022equivariant}, namely atomic stability (the fraction
% of atoms that have precisely the right valency) and molecular stability (the fraction of generated molecules for which all atoms are stable), and the MMD distances for bond length, bond angle, and dihedral angle distributions proposed by~\citet{luo2022an}. 
% The results show that our model outperforms or approximates the performance with these diffusion-based methods which are of high computation complexity.
% These results highlight the model's proficiency in generating molecules with realistic and stable 3D conformations.

\begin{table}[]
\begin{center}
% \vspace{-15 pt}
\caption{Results on 3D conformation generation quality evaluation, with metrics for molecular and atomic stability, as well as MMD for bond, angle, and dihedral angles. }\label{table:results_3d}
\resizebox{0.5\textwidth}{!}{
\begin{tabular}{lccc|cccc}\toprule
&\multicolumn{3}{c}{Geom-Drugs} &\multicolumn{3}{c}{CrossDocked2020} \\\cmidrule{2-7}
&EDM &HierDiff &Ours &EDM &HierDiff &Ours \\\midrule
Mol. Sta. (\%) &0.1 &\textbf{31.7} &{25.6} &0.2 &{35.7} &\textbf{50.9} \\
Atom Sta. (\%) &{81.2} &\textbf{88.4} &79.9 &73.1 &{83.1} &\textbf{87.3} \\ 
\midrule
Bond MMD &1.034 &{0.989} &\textbf{0.712} &{0.487} &0.580 &\textbf{0.341} \\
Angle MMD &1.200 &\textbf{0.871} &{0.916} &0.562 &\textbf{0.151} &{0.386} \\
Diherdral MMD &\textbf{0.028} &0.082 &{0.070} &\textbf{0.040} &{0.051} &0.094 \\
\bottomrule
\end{tabular}}
\end{center}
\end{table}

% \subsection{Multi-property targeting generation results}
% \label{appendix: multi-prop}
% In this subsection, we apply our model to disentangle multiple target properties and the structure context. Specifically, we conduct experiments with asphericity and QED properties using the GEOM-drug dataset. The results are shown in the following table. We can observe that the performance of multi-property targeting generation slightly dropped. Theoretically, asphericity and QED cannot be completely disentangled, but our model can still achieve a certain degree of explicit control on the multi-property-targeting generation.

% \begin{table}[!htp]\centering
% \caption{Results on multi-property targeting generation with asphericity and QED properties using the GEOM-drug dataset. }\label{tab: multi-target}
% \resizebox{0.4\textwidth}{!}{
% \begin{tabular}{lrrrrr}\toprule
% &\multicolumn{2}{c}{Asphericity} &\multicolumn{2}{c}{QED} \\\cmidrule{2-5}
% &MSE &MAE &MSE &MAE \\\midrule
% EDM &0.594 &0.432 &0.147 &0.319 \\
% HierDiff &0.193 &0.412 &0.136 &0.301 \\
% E3WAE &\textbf{0.106} &\textbf{0.251} &\textbf{0.099} &\textbf{0.269} \\
% \bottomrule
% \end{tabular}}
% \end{table}

\section{Ablation Studies}
\label{appendix: ablation}
\subsection{Ablation Study on Disentangle Representation Learning Framework}

To elucidate the rationale behind our model's design, we incorporate two VAE baselines E3fVAE and E3dVAE, inspired by FactorVAE \cite{kim2018disentangling} and IDEL's objectives from \citet{cheng2020improving}. 

The first baseline, E3fVAE, adds a total correlation term for enhanced disentanglement without compromising reconstruction quality. We conduct experiments adapting the disentangle objective in FactorVAE~\cite{kim2018disentangling} for 3D molecule generation referencing the official implementation on GitHub\footnote{ \url{https://github.com/1Konny/FactorVAE}. The MIT license.}. Specifically, we use a 3-layer EGNN as the discriminator and the training objectives are:
\begin{itemize}
\item The reconstruction loss for 3D molecule graphs: $\mathcal{L}_{Recon}$
\item The KL divergence term for VAE $\mathcal{L}_{\text{KL}} = \mathbb{D}_{\text{KL}}\left( Q_\theta(\mathbf{z}|\mathbf{x}) \| P_{\mathbf{z}} \right)$, where $\mathbf{z} = \text{concat}(\mathbf{z}_p, \mathbf{z}_s)$
\item The total correlation term
\item The discriminator loss to approximate the density ratio in KL term
\end{itemize}

The second baseline, E3dVAE, focuses on minimizing a mutual information (MI) upper bound, a strategy aimed at achieving disentanglement within the embedding space. For training, the loss comprises of the following components:
\begin{itemize}
    \item The reconstruction loss for 3D molecule graphs: $\mathcal{L}_{Recon}$
    \item The KL divergence term for VAE: $\mathcal{L}_{\text{KL}} = \mathbb{D}_{\text{KL}}\left( Q_\theta(\mathbf{z}|\mathbf{x}) \| P_{\mathbf{z}} \right)$, where $\mathbf{z} = \text{concat}(\mathbf{z}_p, \mathbf{z}_s)$
    \item The reconstruction loss given $\mathbf{z}_s$: $\mathcal{L}_{Recon, s}(\mathcal{G}, \mathcal{D}_{s}(\hat{\mathcal{G}} | \mathbf{z}_s))$, where $\mathcal{D}_{s}$  is an auxiliary decoder to reconstruct the 3D molecule graph $\mathcal{G}$ by the context latent variable $\mathbf{z}_s$ 
    \item The property prediction loss given $\mathbf{z}_{h,p}$: $\mathcal{L}_{prop}$
    \item The sample-based MI upper bound between the latent variables: $\mathcal{L}_{MI}(\mathbf{z}_p, \mathbf{z}_s)$. This requires an approximation of the conditional distribution $p(\mathbf{z}_{h, p} | \mathbf{z}_{h, s})$, which is approximated by a separate variational network.
\end{itemize}

E3WAE, differing from the above VAE-based models, is deterministic and has simpler objectives, which helps avoid typical VAE training challenges such as hyperparameter sensitivity. We perform an ablation study with GEOM-Drug dataset on property-targeting generation with Asphericity and QED. The results in Table~\ref{table:ablation_model} reveal a performance drop in E3dVAE compared to E3WAE. Considering this and the architectural simplicity of E3WAE, we opt for a WAE-based framework, E3WAE, as our preferred model.

\begin{table}[ht]
\centering
\caption{Ablation study on the architecture choice.
}\label{table:ablation_model}
\resizebox{0.45\textwidth}{!}{
\begin{tabular}{lcccccc}\toprule
&\multicolumn{2}{c}{Asphericity} &\multicolumn{2}{c}{QED} &\multirow{2}{*}{\# Para.} \\\cmidrule{2-5}
&MSE &MAE &MSE &MAE & \\\midrule
E3fVAE &0.143 &0.381 &0.112 &0.306 &3.2M \\
E3dVAE &0.101 &0.260 &0.098 &0.257 &2.4M \\
E3WAE &\textbf{0.095} &\textbf{0.246} &\textbf{0.072} &\textbf{0.221} &\textbf{2.1M} \\
\bottomrule
\end{tabular}}
\vspace{-5pt}
\end{table}

\subsection{Ablation Study on the Proposed Coordinate Reconstruction Loss}
\begin{figure}[h]
     \centering
     \subfloat[Training Set] 
     {\includegraphics[width=0.35\textwidth]{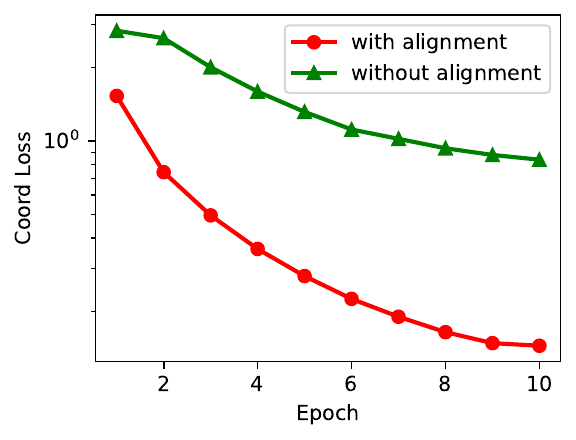}\label{fig:train_pos_loss}}
     \subfloat[Validation Set] 
     {\includegraphics[width=0.35\textwidth]{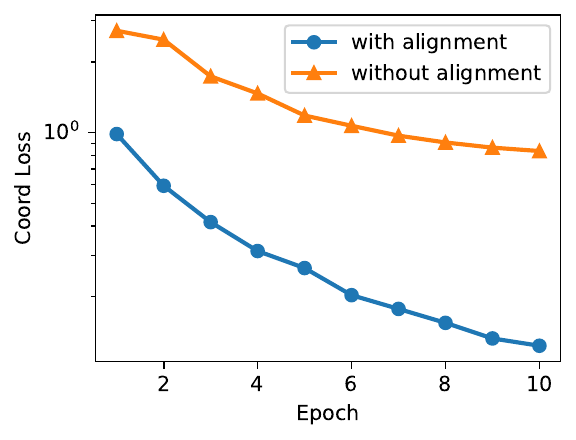}\label{fig:eval_pos_loss}}
    \caption{ 
    % \protect\subref{fig:train_pos_loss}
    Comparison of training coordinate loss curves: proposed coordinate loss vs. original log-MSE loss without structural alignment. }
    \label{fig:coor_loss}
\end{figure}

In this ablation study, we assess the impact of the proposed coordinate reconstruction loss on model training. Figure~\ref{fig:coor_loss} shows the loss curves on training and validation set: with and without the proposed alignment strategy. This indicate that incorporating the alignment into the coordinate loss results in expedited convergence and a lower convergence plateau. This suggests that the alignment component in the loss function plays a critical role in guiding the model to a more accurate and stable solution. Based on these observations, we opt for the proposed coordinate loss in our model to leverage its benefits for more efficient training and improved model performance on validation data.

\section{Visualization of Generated Molecules}
\label{appendix: vis}
\begin{figure}[h]
    \centering
    \includegraphics[width=0.5\textwidth]{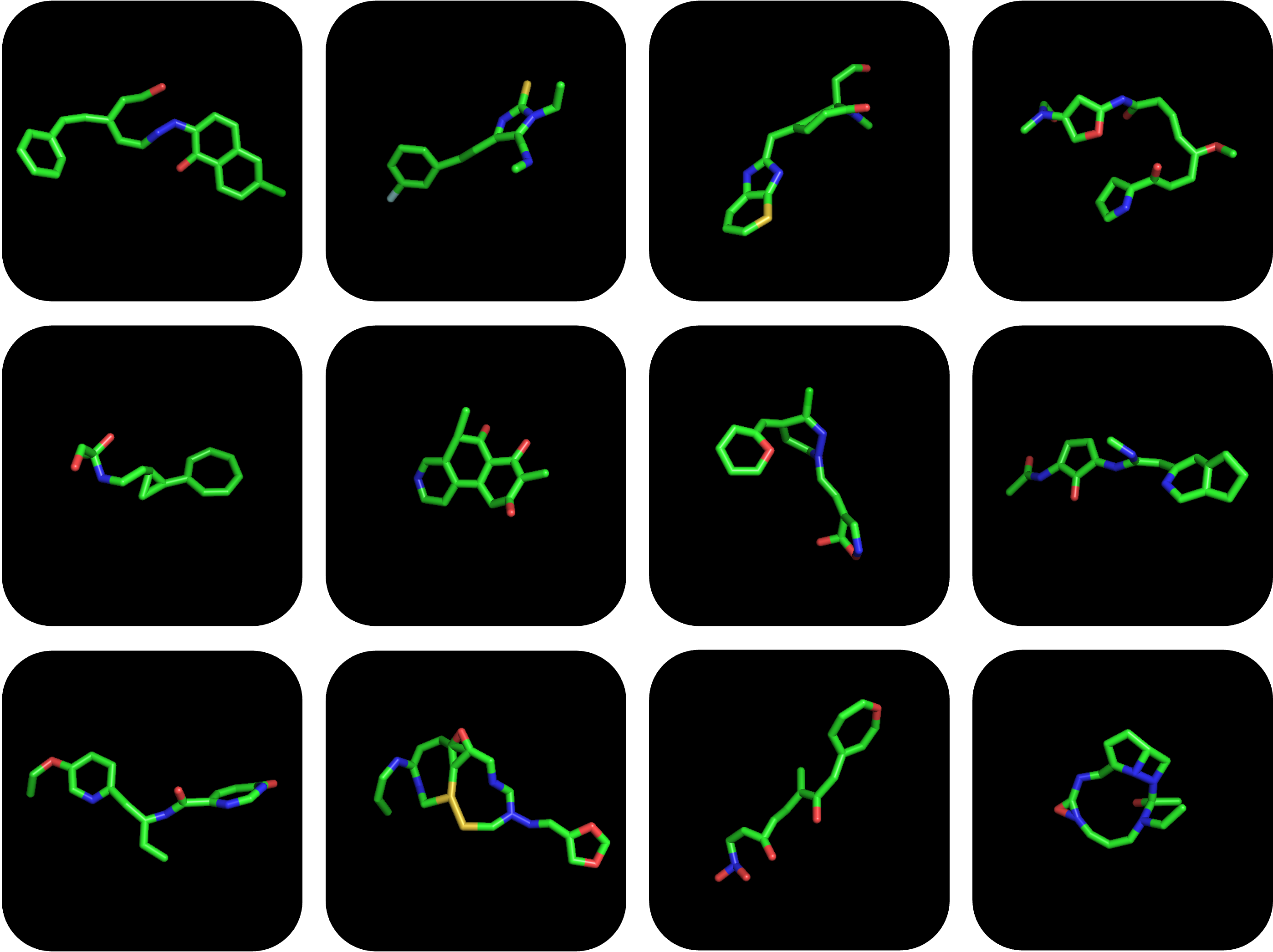}
    \caption{Visualized 3D conformations generated by EDM.}
    \label{fig:vis_edm}
\end{figure}

\begin{figure}[h]
    \centering
    \includegraphics[width=0.5\textwidth]{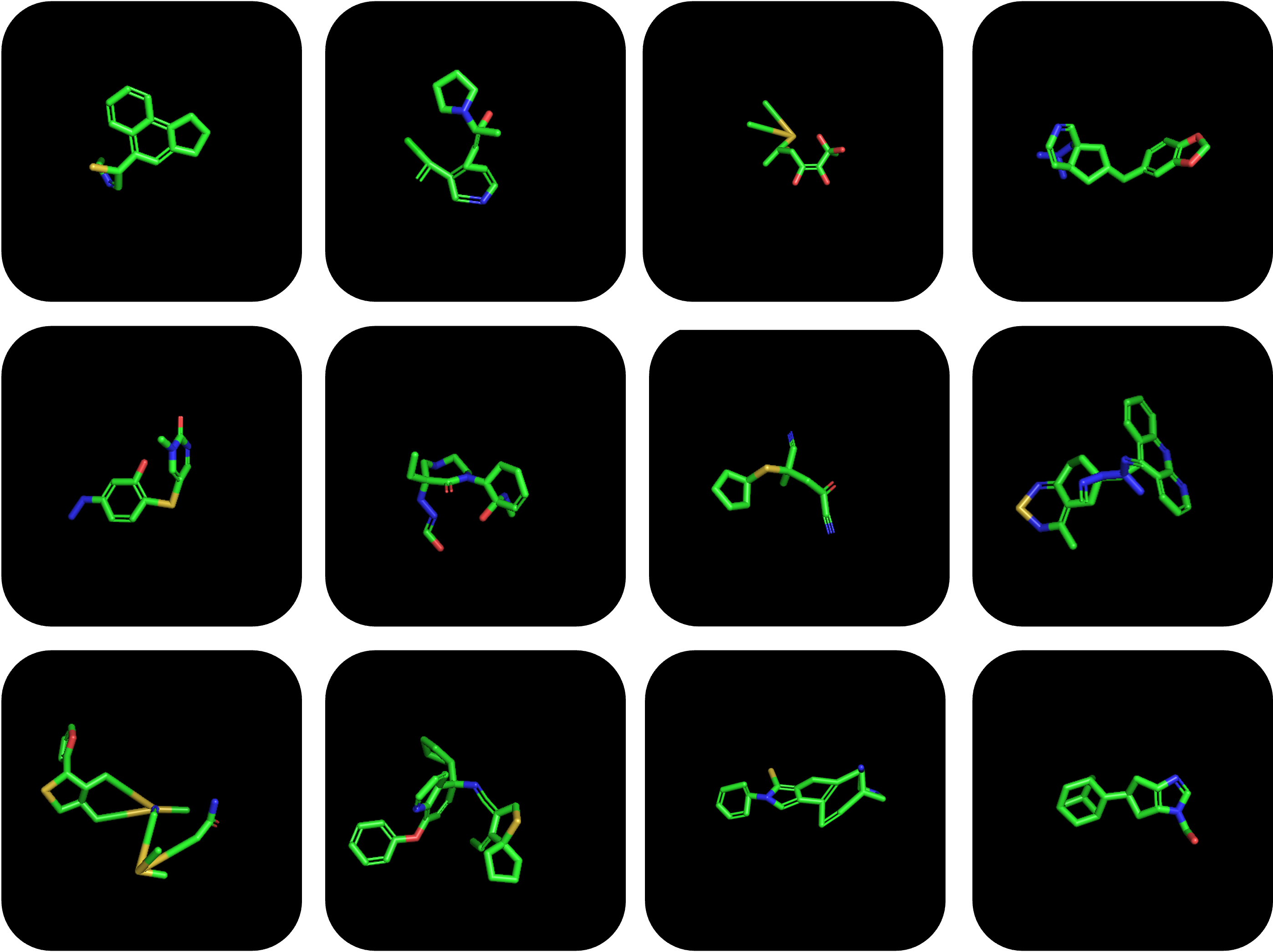}
    \caption{Visualized 3D conformations generated by HierDiff.}
    \label{fig:vis_hierdiff}
\end{figure}

\begin{figure}[h]
    \centering
    \includegraphics[width=0.5\textwidth]{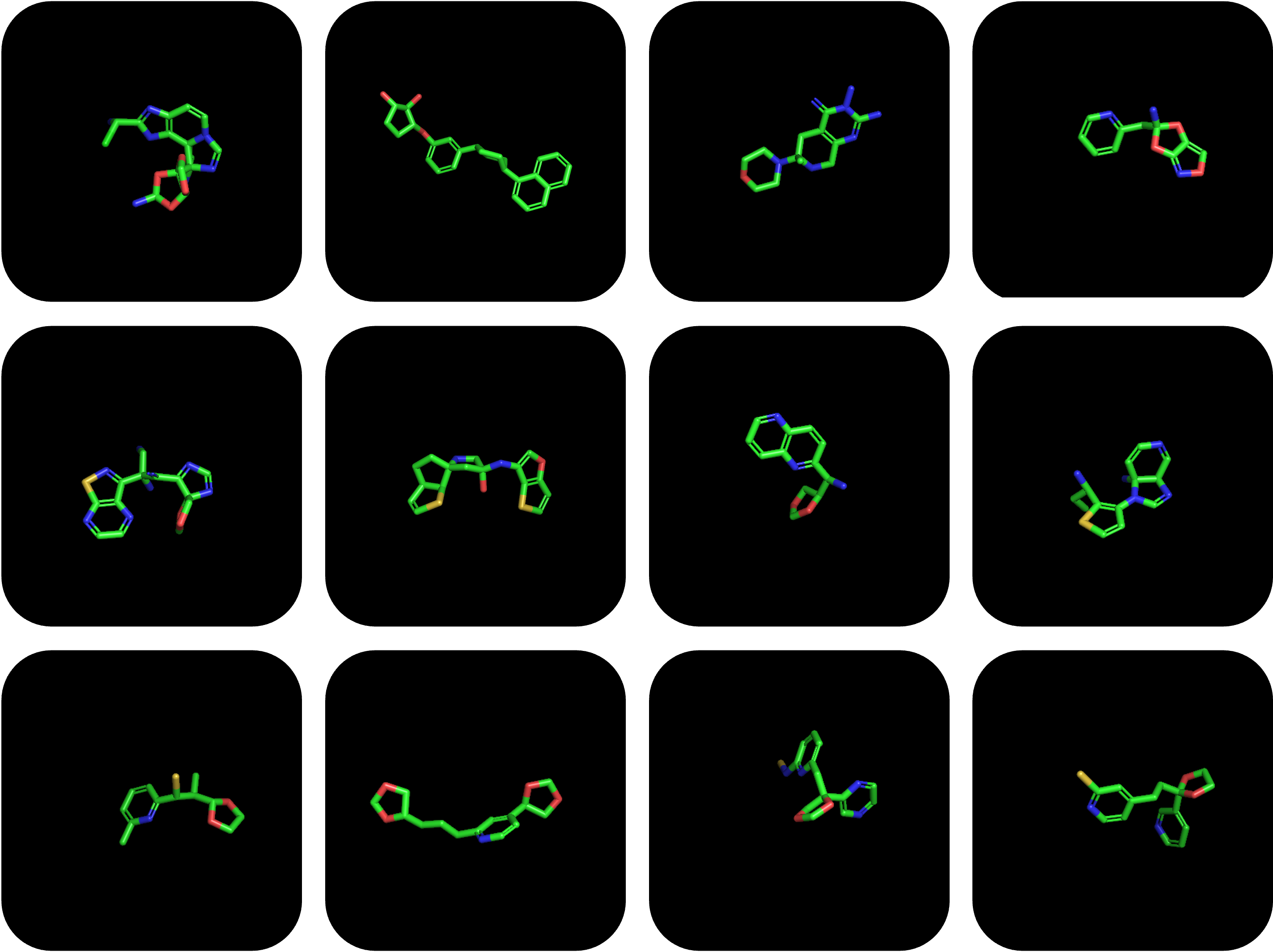}
    \caption{Visualized 3D conformations generated by E3WAE.}
    \label{fig:vis_ours}
\end{figure}

The visualizations of generated molecules for asphericity-guided generation tasks are shown in Figure~\ref{fig:vis_edm} for EDM, Figure~\ref{fig:vis_hierdiff} for HierDiff, and Figure~\ref{fig:vis_ours} for our proposed E3WAE, respectively.
It's worth noting that while EDM may exhibit unstable or broken substructures, such as unrealistic rings, and HierDiff could produce unusual connections, our model consistently generates high-quality molecules with more stable substructures.

% \section{The reconstruction algorithm}

% \paragraph{$\text{NodeTypes}(\mathbf{z}_{h}$, $\mathbf{z}_{v})$}
% \paragraph{$\text{PredictEdge}(f, \mathcal{G})$}
% \paragraph{$\text{PredictCoords}(i)$}

\section{Limitations} \label{limitation}

While our model provides explicit control over molecular properties and structural contexts, the interpretability of the context latent vectors remains a challenge. Supervision through reliable labeled data and human domain prior are required for disentangled representation learning model design. On the other hand, the generalization ability of our model lacks exploration. More consistent and comprehensive evaluations are needed to confirm the robustness and reliability of our approach in specific drug design scenarios and domains.

\section{Broader Impacts} \label{impacts}
This paper presents work whose goal is to advance the field of Machine Learning. There are many potential societal consequences of our work, none which we feel must be specifically highlighted here.

% \begin{figure}[h]
%      \centering
%      \subfloat[EDM] 
%      {\includegraphics[width=0.4\textwidth]{figs/edm_asp_new.png}\label{fig:vis_edm}}
%      \subfloat[HierDiff] 
%      {\includegraphics[width=0.4\textwidth]{figs/hierdiff_asp_new.png}\label{fig:vis_hierdiff}}
%     \caption{ 
%     % \protect\subref{fig:train_pos_loss}
%     Visualized 3D conformations generated by baseline methods. }
%     \label{fig:coor_loss}
% \end{figure}


%%%%%%%%%%%%%%%%%%%%%%%%%%%%%%%%%%%%%%%%%%%%%%%%%%%%%%%%%%%%%%%%%%%%%%%%%%%%%%%
%%%%%%%%%%%%%%%%%%%%%%%%%%%%%%%%%%%%%%%%%%%%%%%%%%%%%%%%%%%%%%%%%%%%%%%%%%%%%%%
\clearpage

\end{document}